\documentclass[10pt]{article}
\usepackage{parskip}

\usepackage{amsmath}
\usepackage{amssymb}

\usepackage{newfloat}
\DeclareFloatingEnvironment[name={Supplementary Figure}]{suppfigure}

\usepackage[frozencache,cachedir=minted-cache]{minted}

\newcommand{\ocaml}[1]{\mintinline{ocaml}{#1}}
\newcommand{\code}[1]{\texttt{#1}}
\newcommand{\defn}[1]{\textbf{#1}}
\usepackage[pdfborder={0 0 0},colorlinks=true]{hyperref}
\hypersetup{urlcolor=RubineRed,linkcolor=BlueViolet,citecolor=ForestGreen}%
\usepackage{cleveref}
\def\eg{\emph{e.g}\onedot} 
\usepackage{subcaption}
\newcommand{\solved}[1]{
    \FPdiv\result{#1}{4}
    \FPround\roundedresult{\result}{1}
    \SI{\roundedresult}{\percent}
}
\usepackage{wrapfig}

\crefname{suppfigure}{supplementary figure}{supplementary figures}
\Crefname{suppfigure}{Supplementary Figure}{Supplementary Figures}

\usepackage[ruled,vlined]{algorithm2e}
\usepackage{graphicx}

\usepackage{graphicx}

\usepackage{cite}

\usepackage{color} 
\usepackage{booktabs}

\usepackage{setspace} 
\usepackage{multirow}
\usepackage{color}
\usepackage[usenames,dvipsnames]{xcolor}

\usepackage[labelfont=bf,labelsep=period,justification=raggedright]{caption}

\makeatletter
\renewcommand{\@biblabel}[1]{\quad#1.}
\makeatother

\makeatletter
\DeclareRobustCommand\onedot{\futurelet\@let@token\@onedot}
\def\@onedot{\ifx\@let@token.\else.\null\fi\xspace}

\def\eg{\emph{e.g}\onedot}

\def\etal{\emph{et al}\onedot}
\makeatother

\date{}

\title{\textbf{Neural networks for abstraction and reasoning:
\\\vspace{-2mm}{Towards broad generalization in machines}}}

\author{Mikel Bober-Irizar\\ {\tt\small mikel@mxbi.net}\and Soumya Banerjee\\ {\tt\small sb2333@cam.ac.uk}}

\date{%
University of Cambridge, UK}

\begin{document}

\maketitle

\section*{Abstract}

For half a century, artificial intelligence research has attempted to reproduce the human qualities of abstraction and reasoning - creating computer systems that can learn new concepts from a minimal set of examples, in settings where humans find this easy. While specific neural networks are able to solve an impressive range of problems, \textbf{broad generalization} to situations outside their training data has proved elusive. %

In this work, we look at several novel approaches for solving the Abstraction \& Reasoning Corpus (ARC), a dataset of abstract visual reasoning tasks introduced to test algorithms on broad generalization. Despite three international competitions with \$100,000 in prizes, the best algorithms still fail to solve a majority of ARC tasks and rely on complex hand-crafted rules, without using machine learning at all. We revisit whether recent advances in neural networks allow progress on this task, or whether an entirely different class of models are required.

First, we adapt the \textbf{DreamCoder} neurosymbolic reasoning solver to ARC. DreamCoder automatically writes programs in a bespoke domain-specific language to perform reasoning, using a neural network to mimic human intuition. We present the \textbf{Perceptual Abstraction and Reasoning Language} (PeARL) language, which allows DreamCoder to solve ARC tasks, and propose a new recognition model that allows us to significantly improve on the previous best implementation.

We also propose a new encoding and augmentation scheme that allows large language models (LLMs) to solve ARC tasks, and find that the largest models can solve some ARC tasks. LLMs are able to solve a different group of problems to state-of-the-art solvers, and provide an interesting way to complement other approaches.

We perform an ensemble analysis, combining models to achieve better results than any system alone. Finally, we publish the \code{arckit} Python library to make future research on ARC easier. %

\section{Introduction}
\label{section_intro}

For the past fifty years, researchers in the field of artificial intelligence (AI) have been striving to replicate human abilities of abstraction and reasoning - developing computer systems that can learn new concepts from a small number of examples, something that humans find relatively easy~\cite{bongard}. While most recent AI  research has focused on \textit{narrow intelligence} (solving specific tasks given large amounts of data), we revisit whether new advances can allow computers to extrapolate to new concepts rather than merely interpolate.

This concept has been referred to as \textit{broad generalization}\cite{arc}: humans do this through abstraction and reasoning; drawing analogies to previous situations and thinking logically. While AI systems such as neural networks excel at a wide range of tasks, from protein folding~\cite{alphafold} to playing Go~\cite{alphago}, their inability to broadly generalize has precluded deployments in the real world, such as in self-driving cars~\cite{selfdriving, badgeneralization}. To enable safer AI, we must understand how to build systems that can reason in unusual situations.

To systematically build and evaluate computer systems that can solve abstract reasoning problems in a human-like intelligent way, we turn to a concrete benchmark. In 2019, the Abstraction and Reasoning Corpus (ARC) was introduced, as an attempt to codify a benchmark of intelligence~\cite{arc} - a sort of `IQ Test' for AI. The ARC contains a series of human-designed tasks on grids, which require learning some transformation from a small number of demonstrations. Despite three international competitions with over \$100,000 in prize money, performance on ARC has proved elusive, with no single system surpassing 50\% accuracy on the ARC-Easy dataset or 20\% on the test set~\cite{arc_kaggle}. %

With only a handful of training examples per ARC task, and $10^{900}$ possible answers (of which exactly one gains credit), traditional machine learning (ML) methods that require large datasets have so far been unable to make progress. Current state-of-the-art approaches to ARC make use of complex hand-crafted algorithms based on brute force search, without harnessing any ML~\cite{icecuber_blog}. This work looks at whether novel machine learning systems that focus on abstraction and reasoning can achieve broad generalization, using ARC as a focal point.%

We investigate two new approaches to ARC, focusing on novel ways to incorporate neural networks to build better abstraction and reasoning solvers. Specifically:

\begin{itemize}
\item We adapt the DreamCoder algorithm, a recent state-of-the-art algorithm for program induction, to solve ARC tasks. DreamCoder writes programs in a domain-specific language; we design the \textbf{Perceptual Abstraction \& Reasoning Language} (PeARL) pure functional language for this purpose. Our DreamCoder implementation solves $3\times$ more tasks than existing work~\cite{alford_thesis}.
\item We introduce a new framework for solving ARC tasks using large-language models (LLMs), transforming these visual tasks into a textual domain. A detailed evaluation of three model classes~\cite{gpt3, gpt4, llama} shows that LLMs can achieve competitive performance with human-crafted systems with the right augmentation and domain transformation.
\item We build an ensemble of multiple ARC solvers, accounting for heterogeneous performance. This ensemble approach leverages multiple complimentary techniques and represents an improvement over the current state-of-the-art~\cite{icecuber_blog}.
\item We publish an open-source Python library for working with ARC to stimulate work in this field.
\end{itemize}

\section{Background}

Since the very first research on artificial intelligence, analogy-making has been considered central to the notion of intelligence. When presented with novel situations (for example; opening a new type of door, or conversing about a new topic), humans effortlessly solve these situations by creating analogies to previous experiences and concepts.

In 1967, Mikhail Bongard made one of the first attempts at identifying this notion of analogy-making in his book \textit{Pattern Recognition} \cite{bongard}. He noted how scientists such as Alan Turing have long posited the concept of a thinking machine; but while machines can be built to solve specific tasks (such as solving quadratic equations or playing chess), no progress had been made to imitate or even understand the ability of humans to adapt to new situations. Bongard goes on to suggest that \textit{pattern recognition}, the ability to recognise situations into \textit{objects} and \textit{classes of objects} (concepts), is central to the abilities of human intelligence.

Bongard introduced a set of problems (now known as Bongard Problems) \cite{bongard}, where two sets of shapes are presented, and the task is to identify the common factor that differentiates them. Note that every Bongard problem has a unique transformation, and thus one cannot simply train a classification model to identify a set of transformations - the system must be able to `invent' them. This rules out classes of models used to solve most AI tasks, which rely on large amounts of training data to spot patterns.

\begin{figure}[h]

    \hspace{-0.05\textwidth}\makebox[1.1\textwidth][c]{
        \begin{subfigure}{0.33\textwidth}
        \fbox{\includegraphics[width=\textwidth]{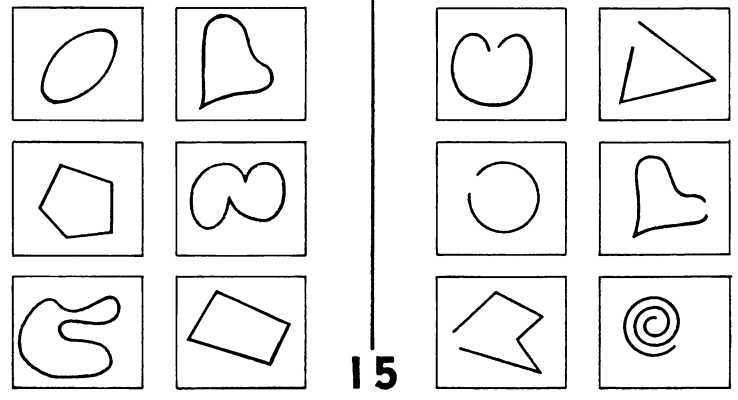}}
        \end{subfigure}
        \hspace{0.02\textwidth}
        \begin{subfigure}{0.33\textwidth}
        \fbox{\includegraphics[width=\textwidth]{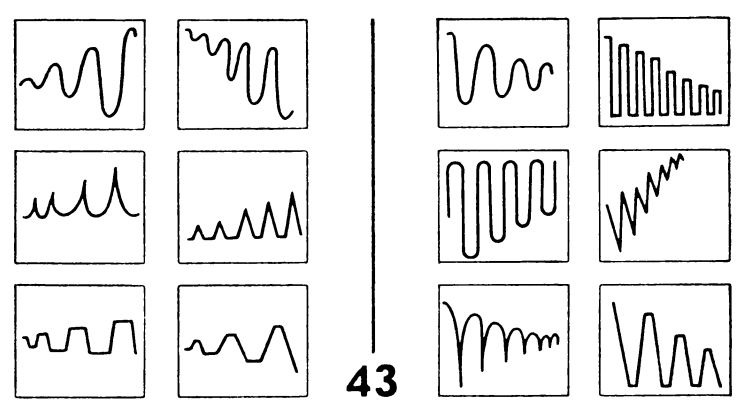}}
        \end{subfigure}
        \hspace{0.02\textwidth}
        \begin{subfigure}{0.33\textwidth}
        \fbox{\includegraphics[width=\textwidth]{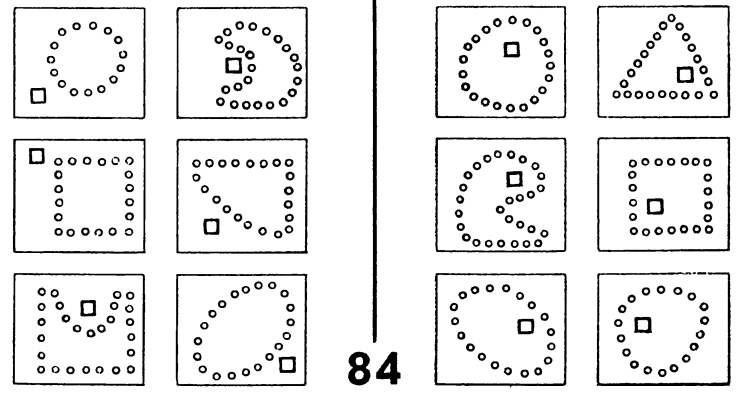}}
        \end{subfigure}
    }
    \caption{Three of Bongard's original problems from 1967. The task is to identify the difference between the two sets, where each problem encapsulates a different concept. {\bf (15)} Set A contains closed shapes while Set B contains open ones. {\bf (43)} Set A contains waves of increasing amplitude; set B decreasing amplitude. {\bf (84)} Set A has a square inside the perimeter defined by connecting the dots; set B outside. Reproduced from~\cite{bongard}.}

    \label{fig:bongard}
\end{figure}

In Douglas Hofstadter's seminal book, \textit{Gödel, Escher, Bach} \cite{geb}, he writes that \textit{``the skill of solving Bongard problems lies very close to the core of `pure' intelligence, if there is such a thing''} - this work popularised the Bongard problems and proposed ideas for solving them that are still relevant today.

In the past 50 years, much work has been done to build on these problems. The original collection has been extended to almost 400 Bongard problems by a variety of authors~\cite{bongard_index}. Many attempts have been made to try and solve Bongard problems computationally, including using neural networks; however, results are still very limited ~\cite{bongard_solvers, bongard_dl, bongard_dsl, phaeaco}. 

Beyond the difficulties of the task itself, Bongard problems are not well-suited to computer evaluation (discussed further in \Cref{sec:bongard_comparison}), so new benchmarks have been proposed in recent years that build on Bongard's ideas. One such benchmark is the Abstraction and Reasoning Corpus.

\subsection{The Abstraction and Reasoning Corpus}

\begin{figure}[b]
    \includegraphics[width=\linewidth]{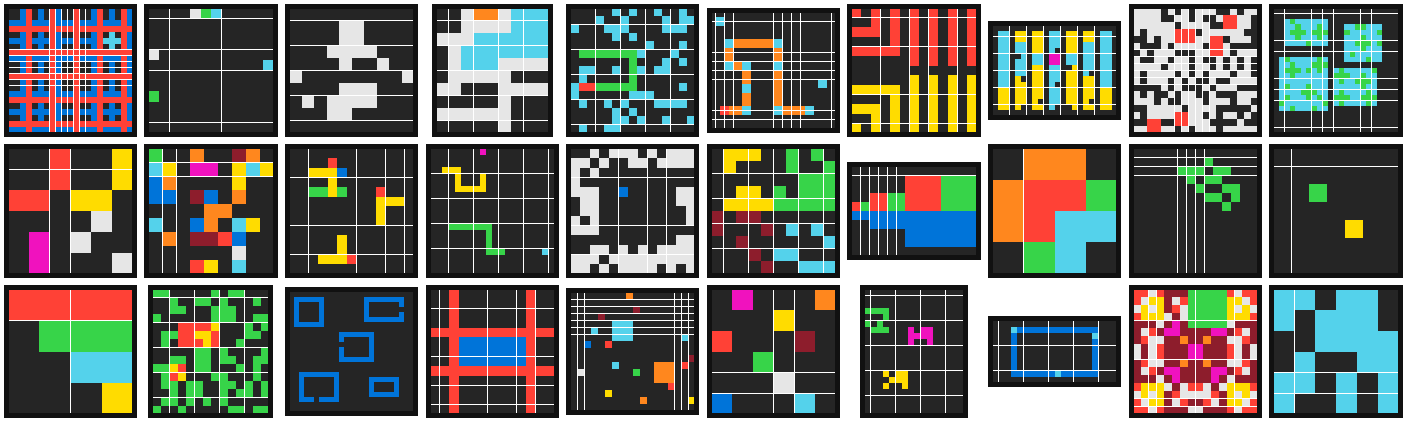}
    \caption{The ARC dataset contains 900 handcrafted abstraction and reasoning tasks based on colourful grids. Each task tests new abilities which must be inferred from few examples.}
    \label{fig:overviewarc}
\end{figure}

In 2019, Chollet published \textit{On the Measure of Intelligence}~\cite{arc}, discussing past and future approaches to general artificial intelligence. Chollet notes that while excellent progress has been made in solving specific tasks to approach or surpass human-level (such as detecting cats and playing Go), these models generally require a huge amount of training and are limited to performing well on situations that they were trained on. The failure of neural network models to perform when extrapolating outside the training data has been widely explored~\cite{cnn-generalization}.

In contrast, humans have an outstanding ability to solve tasks in highly novel situations with little training data, or indeed tasks that no human has ever solved before. 

While machine learning models are often claimed to `generalize', Chollet defines three types of generalization: local generalization, where a system can respond to new examples within an existing domain (\eg an image classifier generalizing to a test set); broad generalization, where a system adapts across a wider set of situations including examples which the system's creator could not have foreseen; and extreme generalization, ``adaptation to unknown unknowns across an unknown range of tasks and domains''~\cite{arc}.%

Capturing and comparing a systems' abilities to perform these sorts of broad generalization tasks is inherently difficult. Despite many decades of research, there is no consensus on how to measure intelligence \cite{definitions_intelligence}, although the extensive field of psychometric testing has proposed many competing tests for humans.

The Abstraction and Reasoning Corpus (ARC), introduced by Chollet in~\cite{arc}, attempts to provide a benchmark for broad generalization. By formalising a concrete way to measure generalization ability, the hope is to foster progress in much the same way as ImageNet transformed image classification~\cite{imagenet_transform}.

\begin{figure}[!b]
    \centering
    \includegraphics[width=\linewidth]{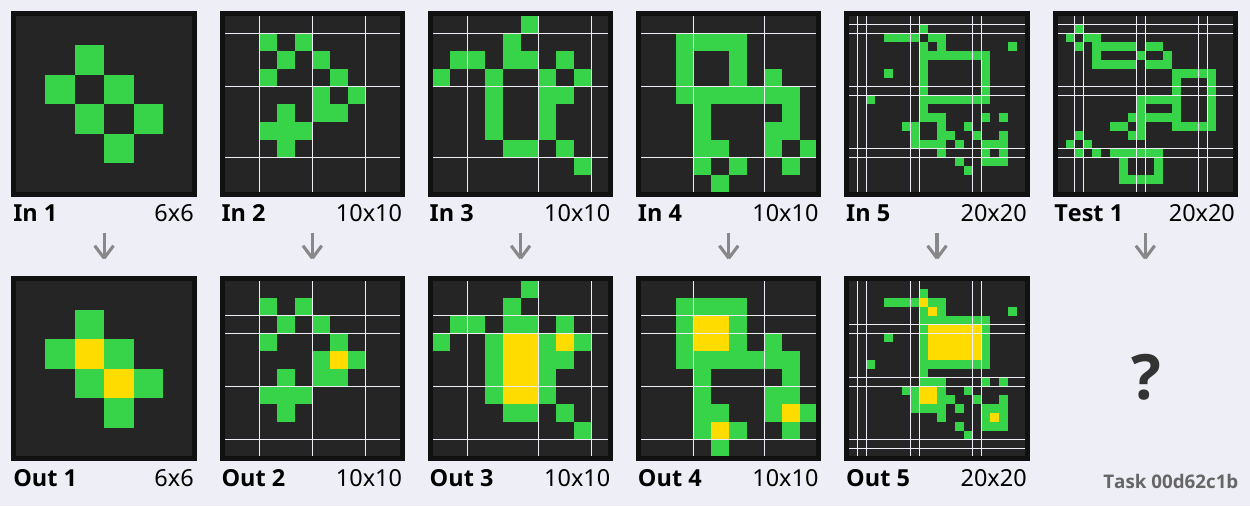}
    
    \vspace{2mm}
    
    \includegraphics[width=\linewidth]{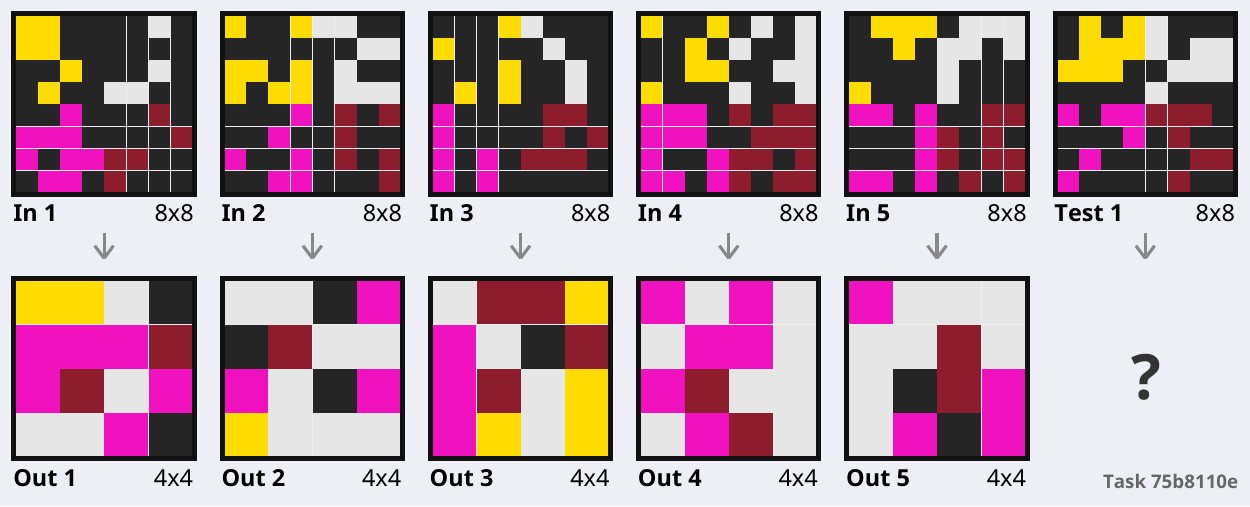}

    \vspace{2mm}

    \includegraphics[width=0.49\linewidth]{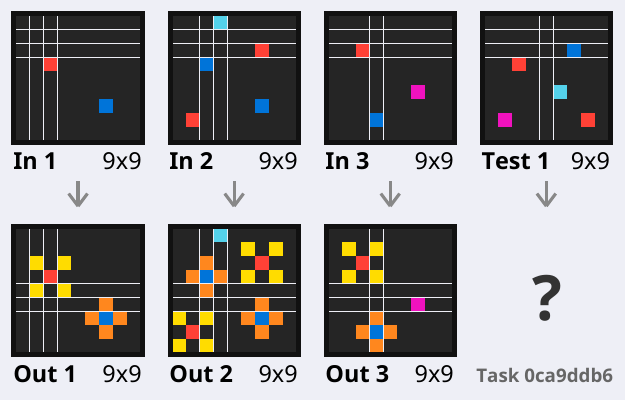}
    \hspace{0.004\linewidth}
    \includegraphics[width=0.49\linewidth]{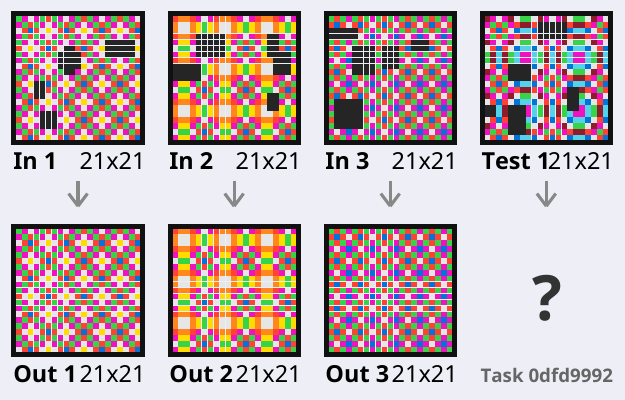}
    \caption{Four examples of tasks from the ARC-Easy dataset. Each task requires recognising new patterns and concepts from the training pairs (In/Out) to predict the output of the test examples from the inputs. The tasks have been hand-designed to test a different set of concepts. %
    }

    \label{fig:exampletask}
\end{figure}

The ARC dataset consists of 900 hand-crafted tasks, each requiring a solver to perform abstract reasoning. In each task, the solver is first presented with some input grids (usually 3-5) and a corresponding set of output grids. Each grid contains pixels of one of 10 colours, represented by integers 0-9 and with a black `background'. The grid size varies between each task and, indeed, within a task. Figures \ref{fig:overviewarc} and \ref{fig:exampletask} show some example ARC tasks.

Each task represents some common transformation from the input to output grids. A system must reason about the differences in training pairs and abstract a transformation to be applied to new input grids to produce output grids. The system is then presented with one or more test input grids, for which the system can provide up to three predictions. A task is considered solved if any of the three predictions are identical to the correct answer - no partial credit is given for close answers.

The dataset is split into three subsets: a \textit{training set}, \textit{evalution set} and \textit{test set}. The training and evaluation sets each contain 400 unique tasks, and the evaluation set contains tasks that are harder than the training set. The private test set contains a further 100 tasks, which are not publicly available: to evaluate a system on the test set, a researcher must submit code to be executed on an offline system with significant resource constraints; as a result we focus on the first two datasets.

Notably, algorithms evaluated in this work use unsupervised learning and do not train on labelled data, meaning we use both datasets exclusively for evaluation. As a result, we refer in this work to these datasets as \textit{ARC-Easy} and \textit{ARC-Hard} respectively.

\subsection{Core Knowledge}
\label{sec:core_knowledge}

Each task is designed to take advantage of some of four \textit{Core Knowledge priors}~\cite{arc}; by defining these explicitly, ARC tries to reduce reliance on whether an algorithm has sufficient `acquired knowledge' and focus purely on reasoning abilities.

\begin{enumerate}
    \item \textbf{Objectness priors: }Handling objects and their interactions. An algorithm must be able to segment the grid into objects based on space and colour (while accounting for noise and occlusion). Physical contact between objects is a common theme (e.g., `gravity' transformations and objects growing until they hit other objects).
    \item \textbf{Goal-directness prior: }Many tasks involve a general notion of `intentionality', finding the simplest solution to some `problem' (e.g.,  drawing the shortest path through a maze rather than a longer one).
    \item \textbf{Numbers and Counting priors: }Some tasks involve counting and basic arithmetic (such as addition and subtraction on numbers below 10), as well as basic set manipulation (such as sorting objects based on some attribute such as size).
    \item \textbf{Basic Geometry and Topology priors: }Many ARC tasks rely on geometric transformations, such as translations, rotations, shape scaling, copying objects, and drawing lines.
\end{enumerate}

Some examples of these tasks are also shown in Supplementary Figure \ref{fig:taxonomy}.

Like Bongard problems, each ARC task is essentially a \textit{few-shot} learning problem. The high dimensionality of the output means that training traditional ML methods on the input/output pairs is impossible; only one of the $10^{900}$ possible output grids gains credit, with as few as three training examples.

Despite the incredibly challenging nature of the ML problem, \textit{an average human can solve a majority of the tasks in ARC}; this highlights our ability to perform broad generalization in a way that today's ML systems cannot, and highlights a significant gap in current AI systems. %

\subsection{Comparison with Bongard problems}
\label{sec:bongard_comparison}

While Bongard problems and ARC are designed to test a system or human's ability to perform inductive reasoning in the presence of few examples, there are key differences. ARC was designed for a modern machine learning paradigm. This makes it more amenable to both designing algorithms to solve it and robustly evaluating those algorithms, compared to Bongard problems and classical psychometric tests designed for humans (such as Raven's Progressive Matrices~\cite{raven_progressive}).

\begin{itemize}
\item \textbf{Evaluation:} How would we unambiguously determine if a system correctly identified the abstract transformation in a Bongard problem? Determining whether any analogy description is `correct' could be subjective and time-consuming. %

In ARC, the problem is modified: the system is instead presented with a few examples of a transformation and tasked with applying
the transformation to a new input. The output can then be scored algorithmically (the model was successful if it produces a pixel-perfect output).

\item \textbf{Data size:} While Bongard published 100 of his problems, ARC has 900 total tasks. More tasks means both that learning systems have more training data, and allows for more precise evaluation.%

Additionally, 100 tasks are held back as a private test set: by enforcing that these tasks are unseen, one can truly test `developer-aware' generalization~\cite{arc}.

\item \textbf{Problem format:} Distilling the problem of logical reasoning (rather than interpreting or generating images), ARC presents tasks as coloured pixels on a variable-size grid. This means that systems do not have to rely on a computer-vision shape detector: the problem is distilled as far as possible into one of reasoning.
\end{itemize}

These attributes make ARC an excellent testbed for ongoing research into the ability of systems to perform abstraction and reasoning. Benchmarks have led to immense progress in other areas of AI research, such as the ImageNet image classification challenge. A similar benchmark for abstraction and reasoning may accelerate research into broad generalization in machines.

\subsection{Previous Work}
\label{sec:related_work}

Many attempts have been made to computationally solve ARC, primarily through the Kaggle Abstraction \& Reasoning Challenge hosted in 2019~\cite{arc_kaggle} with a \$20,000 prize pool, where the current state-of-the-art was set by Johan Sokrates Wind (a.k.a. Icecuber)~\cite{icecuber_blog}. Icecuber implements a Domain-Specific Language (DSL) with 142 handcrafted unary functions on grids. At runtime, the functions are greedily composed on the input grids, with the resulting `pieces' stored in a directed acyclic graph (DAG). Finally, a solver combines pieces from the DAG to get as close as possible to matching the training examples. Figure \ref{fig:icecuber} shows an outline of the approach; \textbf{a detailed description is available in Supplementary Materials}.

\begin{figure}[h]
    \hspace{-0.05\textwidth}
    \makebox[1.1\textwidth][c]{
    \includegraphics[width=1.1\linewidth]{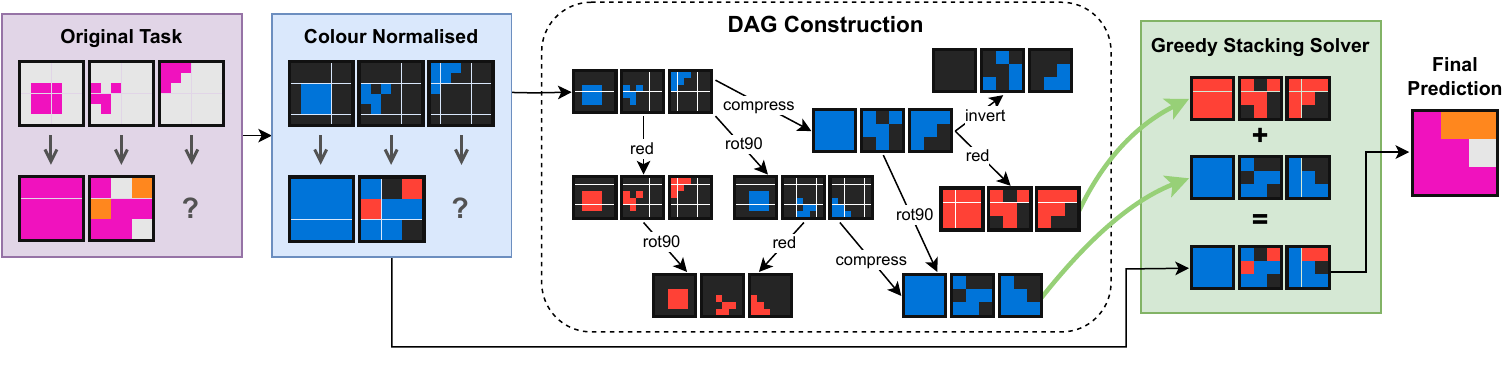}}
    \caption{Logical diagram of the Icecuber algorithm. Input grids for a task are combined to form `pieces'. A brute-force search over unary functions on pieces forms a directed acyclic graph (DAG) of potential outputs, until outputs are found that match the task's outputs. If no exact match is found, a greedy stacker combines DAG nodes to produce the final result, minimising pixel distance to the training outputs.}
    \label{fig:icecuber}
\end{figure}

Xu \etal introduced ARGA (Abstract Reasoning with Graph Abstractions)~\cite{arga}; they extended DSL search by converting ARC grids into an object-graph representation, and operating on these representations instead. Additionally, a series of \textit{constraints} were derived from the training examples to prune the search space, and a `Tabu list' avoided searching primitives with poor recent performance. Overall, this approach achieved performance comparable to Icecuber limited to a small subset of tasks related to object manipulation. %

\enlargethispage{1\baselineskip} 
While ARC was designed to be a machine learning benchmark, the state-of-the-art solutions all rely on entirely handcrafted methods similar to Icecuber. There have been attempts to use ML, but these have been limited to small subsets of the ARC dataset. For example, Golubev \etal solved tasks that rely on \textbf{cropping} by extracting features from grids and training a task-specific decision tree classifier to try and predict cropping coordinates $(x, y, w, h)$ for a task's test example~\cite{golubev_soln, decisiontree}. This approach generalized to solve 7\% of private test set tasks. 

In 2021, Alford \etal~\cite{alford_thesis, dreaming_arc} explored the idea of applying the neurosymbolic solver DreamCoder~\cite{ellis_dreamcoder_2020} to the ARC dataset, considered much more challenging than previous DreamCoder applications. They selected a subset of 36 ARC-Easy tasks involving symmetry and basic geometric operations (such as rotations, flips and crops), and provided a DSL with 5 basic primitives on grids. They found that DreamCoder could solve some of the tasks and also successfully created new primitives. However, this solution was limited to a small subset of the ARC-Easy dataset, and did not use neural-network-guided search unlike the original neurosymbolic solver DreamCoder paper~\cite{ellis_dreamcoder_2020}.

In the next section, we review the DreamCoder algorithm in detail, and then revisit whether DreamCoder can be extended to solve much more of ARC by harnessing a more powerful DSL, neural networks and other improvements.

\subsection{Neurosymbolic programming with Dreamcoder}

Within AI, the field of inductive programming \cite{ilp} describes algorithms that derive \textit{programs} that explain a series of examples. After 30 years of research, many algorithms and approaches have been proposed~\cite{ilp30} across a wide range of applications, with the research area far from being solved.
In general, inductive programming provides an encouraging research direction for ARC due to its ability to massively prune the search space from ``all possible grids" to just those explainable by a programmatic transformation.

The DreamCoder system~\cite{ellis_dreamcoder_2020}, was a novel approach to inductive programming, which used neural networks to guide its ability to write programs (neurosymbolic programming).

The DreamCoder algorithm can be broken up into multiple phases, and can be thought of as a \textit{wake-sleep} algorithm. During \textit{waking} phase, a generative model writes programs in a domain-specific language (DSL) that attempt to solve tasks. In the two sleeping phases, the programming language is updated to consolidate new information learned in waking, and a separate \textit{recognition model} is trained which learns to guide the search towards promising programs. 

These phases are interleaved in several iterations to allow for self-improvement. The result is that DreamCoder can achieve remarkable performance across several domains, such as list processing, reproducing LOGO drawings, and finding regexes~\cite{ellis_dreamcoder_2020}.

In this section, we provide a detailed explanation of the DreamCoder algorithm, which we later adapt in Section~\ref{sec:dreamcoder}. %

\subsubsection{Waking phase}

DreamCoder, in a general form, attempts to write programs to solve tasks; we must first define what this means. We define a \textit{task} $T(x, y): p$ as a set of examples $(x, y)$ defined by a program $p$ such that $p(x) = y$. For a list processing task, $(x, y)$ could contain pairs of unsorted and sorted lists, with the correct program $p$ being a sorting algorithm. For tasks without input/output pairs (e.g, generating a program that reproduces a drawing), $x$ can be empty.

The goal of DreamCoder is to inductively reason about the examples in $T$ to produce a candidate program $p'$ which matches the examples. It is important that tasks are \textit{verifiable}: one can check programmatically whether a program is the correct solution to a task. In our list processing example, we can compute whether $p'(x) = y$, even if $p$ is not known; this means that we can apply it to real-world problems where the answer is unknown, but can be verified.

DreamCoder is bootstrapped with a Domain-Specific Language known as a library. Just like any programming language, the library contains a set of functions and values, defined within a functional type-system. Making use of the type-system, we can generate a grammar that recursively defines the set of well-typed programs within the language. For example, consider the following toy language:

\begin{center}
\begin{tabular}{rl}
\textcolor{black!60}{\textbf{Types:}} & \code{int, \enskip str} \\
\textbf{\textcolor{black!60}{Values:}} & \code{1, 2, 'a', 'b'} \\
\textbf{\textcolor{black!60}{Functions:}} &      \code{add: int -> int -> int} \\
    &\code{concat: str -> str -> str}
\end{tabular}
\end{center}

In this language, \code{add 1 2} is a well-typed program but \code{add 1 'a'} is not, even though it is syntactically correct. With a type-system, one can constrain the search space of programs to valid ones: a powerful tool since the set of syntactically correct programs is commonly astronomically bigger than the set of well-typed programs.%

The waking phase of DreamCoder makes great use of this grammar: for each task $T$, we enumerate a large number of candidate programs $p' \in L$; for each sampled program, we check if it solves the task. As there are infinite possible programs, we must use a heuristic to decide which programs to sample. DreamCoder uses the \textit{Minimum Description Length} (MDL) principle, by computing the entropy of each program and enumerating the programs with the least entropy first. This heuristic is based on the idea that the shortest program that solves a task is the most likely to be the correct one (often compared to Occam's razor).

On its own, the waking phase can be seen as a brute-force search with a powerful and cleverly-defined search space. %

\subsubsection{Abstraction Sleep}
\label{sec:dreamcoder:abstraction}

The power of the DreamCoder algorithm comes from the two sleep phases, known as \textit{abstraction sleep} and \textit{dreaming sleep}.

Abstraction sleep considers and manipulates the solutions of tasks solved during waking. First, discovered solution programs (\defn{frontiers}) $P$ are inserted into a \textit{version space}; a data structure that efficiently represents possible refactorings of programs. 

The version space represents a large set of programs: for example, the functional program $(+\ 1\ 1)$ can be refactored as $((\lambda\ (x)\ (x\ 1\ 1))\ +)$, and both of these would be included in the version space. All programs within $n=3$ steps of refactoring are considered.

By representing all frontiers along with possible refactorings in the version space, we can pick out common concepts that occur across multiple programs as new \textit{primitives}. These primitives can then be added to our library for the next iteration of waking, adding common concepts to the DSL (learning by analogy to existing tasks). At each iteration, the $k$ most common concepts that cause the most reduction in total description length are compressed into new primitives.

The effect of abstraction sleep is to dramatically reduce the depth of search (at the cost of a slight increase in breadth). In practice, this abstraction learning can be extremely powerful: in a LOGO drawing task, the learned concepts included drawing polygons with $n$ sides and $l$ side length, or drawing a circle with $r$ radius~\cite{ellis_dreamcoder_2020}. The concepts learnt can even include higher-order functions, such as a radial symmetry function that repeats a function multiple times at different angles. At the next iteration of waking, these concepts form longer, more powerful programs than is possible by DSL search on simple operations.

\subsubsection{Dreaming Sleep}
\label{sec:dreamcoder:dreaming}

The final phase of the algorithm is \textit{Dreaming Sleep}. In this phase, we focus on decreasing the breadth of search by intelligently guiding the generative grammar for each task. To do this, we train a neural network \textit{recognition model}, which can directly perform abductive reasoning and infer $T(x, y) \rightarrow p$. There are two challenges to overcome in designing such a model.

First, the space of programs is exponentially large, and tasks are very difficult; even a human will often consider and discard multiple candidate hypotheses before arriving at the correct solution. Hence, a neural network which directly predicts the answer is likely to fail. 

Instead, the neural network is trained to produce a grammar under which the correct solution $p$ has low entropy. Since programs are enumerated in order of entropy (using iterative deepening), this means that a search guided by the recognition network can find the correct solution faster.

\begin{figure}[h]
    \centering
    \includegraphics[width=\linewidth]{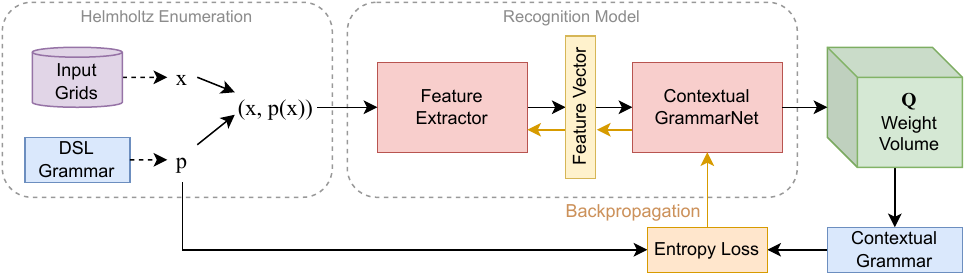}
    \caption{The training process for the recognition model. Helmholtz enumeration generates an unlimited stream of dreamed tasks, and the recognition model attempts to produce a contextual grammar which assigns a low entropy to the correct solution for the dreamt tasks. The model outputs a weight volume $Q$, where $Q_{ijk}(x)$ is the probability of primitive $i$ being the $k$th argument to primitive $j$ for a given task. Backpropagation optimises the neural network end-to-end.}
    \label{fig:recog-training}
\end{figure}

The neural network is made up of a feature extractor which converts a task to a fixed-width feature vector, and a \textit{GrammarNet}, which takes the feature vector and produces a volume $Q$, where $Q_{ijk}(x)$ is the probability of primitive $i$ being the $k$th argument to primitive $j$. From $Q$, we can construct a contextual grammar which assigns likelihoods to programs, and sample from this new grammar during waking. This is shown in Figure \ref{fig:recog-training}. 

The second issue is that of training data: with only 800 tasks available and no labelled solutions ($p$) for any of them, we cannot train a complex neural network on our data. To solve this, DreamCoder uses \textit{dreaming} to generate new training tasks (also called Helmholtz enumeration, inspired by Helmholtz machines~\cite{helmholtz}). To dream up a labelled training task, we randomly sample a program $p^*$ from our existing probabilistic grammar, and sample some input grids $x$ from the empirical distribution of our tasks; we can now construct a new dreamed task $T^*(x, p^*(x)): p^*$. To train the recognition model, a constant stream of dreamed tasks and associated correct programs can be used for backpropagation.%

The role of the recognition model in DreamCoder is very similar to the use of `policy networks' in other works such as \textit{AlphaGo}~\cite{alphago}, which achieved superhuman performance in the board game Go. In AlphaGo, a Monte-Carlo tree search (MCTS) is used to evaluate possible positions on a board, with a policy network suggesting potentially useful moves to evaluate: the neural network acts to dramatically prune the search space and make search feasible. 

In DreamCoder, the enumeration engine acts as the MCTS (evaluating programs matching a recursive grammar), while the recognition model assigns weights to the grammar such that more `promising' programs are evaluated first. \textit{These networks attempt to to mimic a human's ability of intuition}: when a person solves ARC tasks or plays Go, the vast majority of valid operations are immediately discarded as nonsensical.

\section{Methods}

\subsection{Adaptation of DreamCoder}
\label{sec:dreamcoder}

We adapt DreamCoder as an ARC solver, combining the power of DSL search with neural networks. We build on the reference Python/OCaml implementation by Ellis et al.~\cite{dreamcoder_github}, as well as the work of Alford et al.~\cite{alford_github}. To allow DreamCoder to effectively solve ARC tasks, we introduce a new recognition model and domain-specific language (PeARL) for operating on grids, described below. Additional adaptations such as a new evaluation module and multicore search were also implemented, and are open-sourced with this work (\Cref{sec:software}). %
We detail the two primary modifications to the DreamCoder architecture below; \Cref{fig:dreamcoder_overview} also gives a top-level view of the modules modified from the reference DreamCoder implementation.

\subsection{ARC Recognition Model}

A core component of the DreamCoder architecture is the \textbf{recognition model}. With many primitives available to our model, a brute-force search is costly and cannot go sufficiently deep to find all solutions within the search space. The recognition model is a neural network which attempts to \textit{guide} the search, by estimating which programs are most likely to solve a task before enumeration. It is trained during dreaming sleep, and produces a contextual grammar for each task, which assigns a high probability to the correct solution (\Cref{sec:dreamcoder:dreaming}).%

\newcommand{\Tau}{\mathrm{T}}

The recognition model begins with a feature extractor module which converts a task to a fixed-width feature vector. This allows a different feature extractor to be used for different applications (such as LOGO drawings or list processing~\cite{ellis_dreamcoder_2020}). At the same time, a common Grammar Network learns to induce grammars for each task based on extracted features. The end-to-end network can be optimised using gradient descent by making the feature extractor differentiable. %

\begin{figure}[H]
    \centering
    \includegraphics[width=\linewidth]{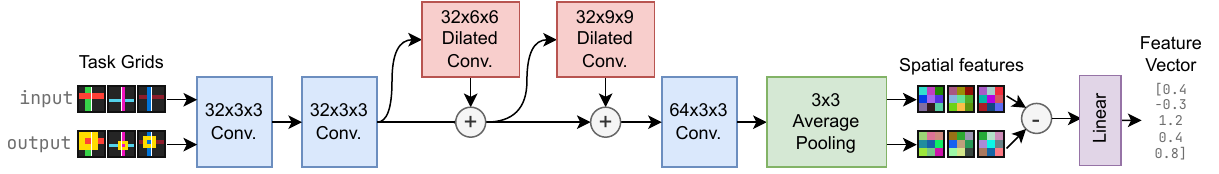}
    \caption{The architecture of the fully-convolutional feature extractor. Each grid within an ARC task is operated on by a series of fully-convolutional layers to produce spacial features; the residual between input and output features is turned into a vector that describes the task. The grid size is maintained as long as possible, and the differential of input and output grids are used as features. Residual dilated convolutions ensure multi-scale context can be used. This architecture allows 1x1 up to 30x30 grids to map to the same vector space, as ARC tasks do not have a fixed grid size.}
    \label{fig:arcnet}
\end{figure}

ARC tasks have a very different format to previous applications of DreamCoder and thus necessitate a new feature extractor design. The 2D image-like nature of our grids means that we look to image recognition networks as a base, but there are several design challenges to overcome. We discuss these in turn. The final selected architecture is shown in \Cref{fig:arcnet}.

First, the highly variable grid size precludes the use of convolutional neural networks (CNNs), which rely on a fixed-size image input, or at least a minimum-size input when adaptive or global pooling layers are used~\cite{adaptive_pooling, global_pooling}. In our case, we need a network that can effectively operate down to $1 \times 1$ grids.

One option is to pad all grids to a sufficiently large fixed size, such as $30 \times 30$. However, this risks hurting performance on tiny grids when most of the image is padding (most inputs are much smaller than $30 \times 30$). Our network is therefore inspired by Fully Convolutional Networks (FCNs)~\cite{fully_convolutional}, which remove these limitations. We use a series of convolutional layers; each layer has internal padding such that its output size equals its input size. 

Instead of downsampling operations usually employed by CNNs and FCNs to extract larger-scale features (which enforce a minimum input size), we use dilated convolutions~\cite{dilated_conv} in later layers of our network. Dilated convolutions leave gaps between sampled pixels; the resulting behaviour is similar to a convolution on a downsampled image. These layers enable multiple scales of contextual information to be incorporated without the parameter explosion associated with large convolutions. The outputs of the dilated convolutions are added to the feature vectors as residuals, meaning the network can choose not to use them (\eg on small grids).

Another unique attribute of ARC is that we have both input and output grids; rather than extracting features from a single grid, the solution to a task depends on the \textit{relationship} between two grids of potentially different sizes. To generate a single set of features for a \textit{task}, we apply the network to each grid followed by an adaptive average-pooling layer to generate two $64\times3\times3$ feature maps $M(y), M(x)$. The \textbf{difference} between these two feature maps $M(y) - M(x)$ is then passed to a linear layer and averaged across training examples to create a single 256-dimension feature vector for an entire task. Using a spatially adaptive layer allows the network to detect when an object has been \textit{translated} between the input and output grid; this was found to improve performance.

\subsubsection{Model training}

Due to the small model size, we perform all training on CPU. The recognition model is trained at each wake-sleep cycle on Helmholtz-sampled tasks for 360 seconds; around 3,000 random tasks. The Adam optimiser~\cite{adam} optimises the sum of two loss functions: the \defn{entropy loss}, which is the overall log-likelihood of the program given the generated grammar, and a \defn{classification loss} that treats the grammar network as an N-classifier where N is the number of primitives and minimises binary cross-entropy. Halfway through training, we anneal the learning rate $10\times$ to improve convergence. 

Initially, while the model converged to a relatively low loss, its predictions were not useful on actual ARC tasks. Upon inspection, the dreamed programs were extremely complex and too ambiguous for even a human to solve the generated tasks. To combat this, the Helmholtz sampling procedure was modified to limit any sampled program to a maximum depth of 3 primitives, dramatically reducing the loss.

\begin{figure}[H]
    \centering
    \includegraphics[width=0.8\linewidth]{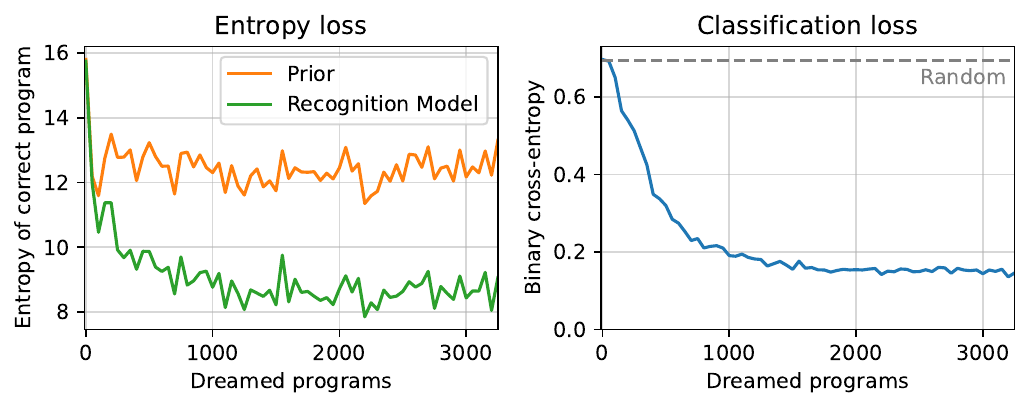}
    \caption{Evaluation metrics for the abstraction sleep phase. Entropy loss gives the average entropy of the solutions generated under the neural-network grammar. In addition, we record the binary cross-entropy loss of classifying which primitives are used in a given task. We see that our recognition model is able to guide the search towards the correct solution; this translates to solving more ARC tasks within a given enumeration time limit.}
    \label{fig:recog_loss}
\end{figure}

\Cref{fig:recog_loss} shows the training behaviour of the recognition model. We see that when neural-network-derived grammars are used, the entropy of discovered solutions (description lengths) is reduced by about 30\% compared to a prior grammar over all tasks. Solutions with lower description lengths take \textbf{exponentially less time to find}, so this improvement is dramatic. We also see that the recognition model learns to classify which primitives might be used to solve a given task. We find empirically that our recognition model reduces the number of programs written before a solution is found by approximately $10\times$.

\subsection{The PeARL language}
\label{sec:methods:pearl}

DreamCoder aims to perform program induction within some well-defined language, using guided search to improve efficiency. In principle, DreamCoder could be used to write programs in any language with recursive grammar (even Python~\cite{python_grammar}).

As discussed in \Cref{sec:error_cases}, searching for all possible Python programs (even a guided one) to discover solutions to ARC tasks would be extremely difficult and inefficient. For this reason, we instead build a \textit{domain-specific language} (DSL) in which a much higher proportion of enumerable programs are likely to be useful. Indeed, current state-of-the-art approaches to ARC mostly involve a DSL in some capacity (see \Cref{sec:related_work}).

Following existing DreamCoder DSLs for other domains, we designed the \textbf{Perceptual Abstraction \& Reasoning Language} (PeARL), a bespoke DSL explicitly designed to represent transformations in ARC tasks. PeARL has two constructs: \textit{types} and \textit{primitives}. Types represent data-types and primitives can represent either a \textit{value} or an n-ary function.%

\pagebreak
We define the following types in PeARL:
\begin{description}
\item [\code{grid}]: A 2D grid of coloured pixels. Each grid contains a size and position, where the position is the original position of that segment (e.g., if the grid has since been cropped).
\item [\code{size}, \code{pos}]: The size or position of a grid, represented by a tuple of integers $(x, y)$. When a subgrid is cropped, it retains a position, which can be used for alignment when recombining subgrids.
\item [\code{colour}]: A pixel colour, represented by an integer 0-9, where 0 is black.
\item [\code{count}]: An integer. This type can be used for arithmetic on grid properties, which is useful for some tasks.
\end{description}

A \textit{valid program} in PeARL is any lambda expression of type \code{grid -> grid} which type-checks. Following existing DreamCoder DSLs, PeARL programs can use any number of primitives, lists, and higher-order functions. The language is entirely defined by a series of Python functions (one for each primitive), where the Python type-annotations implicitly generate the corresponding grammar. For convenience, a \code{DSL} class was implemented, allowing a one-to-one mapping between Python definitions and PeARL, removing additional boilerplate. This enabled rapid prototyping and experimentation.

For example, here is an implementation of a subset of PeARL with two types and two primitives:

\definecolor{mintedbg}{HTML}{F4F4F4}
\begin{minted}[fontsize=\footnotesize, bgcolor=mintedbg]{python}
Colour = NewType("Colour", int) # Colours are ints
class Grid: ... # Implementation of the Grid type is more complex

typemap = {Grid: baseType("grid"), # baseType defines a DreamCoder type
           Colour: baseType("colour")}
dsl = DSL(typemap) # Mapping of python types to DreamCoder types

@dsl.primitive
def topCol(g: Grid) -> Colour:
    return np.argmax(np.bincount(g.grid)[1:])+1

@dsl.primitive
def filterCol(g: Grid, c: Colour) -> Grid:
    filtered = g.grid.copy()
    filtered[filtered != c] = 0
    return g.newgrid(filtered)
\end{minted}
\vspace{-12mm}

The DSL is populated with types \code{grid, colour}, as well as two primitives:

\begin{minted}[bgcolor=mintedbg]{ocaml}
topCol: grid -> colour
filterCol: grid -> colour -> grid
\end{minted}
\vspace{-12mm}

An example of a valid program in this DSL is $\lambda$\code{: filterCol \$0 (topCol \$0)}, which represents the operation ``remove all colours from the grid except the most common one''. The argument \code{\$0} represents the input grid to an ARC task, with the result of the function representing the \textit{output} grid of that task.

A given program \code{f} is \defn{correct} if \code{f(x) = y} for all $(x, y)$ pairs in a task (such programs are known as \defn{frontiers} for a given task). Hence, the \textit{goal of DreamCoder} is to find frontiers for each task, which can be used to generate predictions on the test examples.

\subsection{A framework for large-language models}
\label{sec:llms}

Language models are a class of models designed to statistically model natural language, being trained to predict the next token (\eg words) in a training corpus. Large language models (LLMs) are characterised by their size (containing tens of billions of parameters) and training on a vast corpus of text (usually scraped from the internet). When fine-tuned using reinforcement learning with human feedback, `chat' models can be conditioned to respond to instructions in a conversational format~\cite{rlhf}. %

Since the development of GPT-3 in 2020~\cite{gpt3}, LLMs have gained popularity at a seismic pace, and the ability of LLMs to write code~\cite{codex} or solve word puzzles is impressive. Researchers from Microsoft have stated that GPT-4~\cite{gpt4} ``could reasonably be viewed as an early (yet still incomplete) version of an artificial general intelligence (AGI) system", and found that it could solve some reasoning tasks.%

Further, work from Webb et al. \cite{webb_emergent_2022} has recently suggested that ``large language models such as GPT-3 have acquired an
emergent ability to find zero-shot solutions to a broad range of analogy problems". To demonstrate this, they created a digit matrices problem set inspired by Raven's Progressive Matrices. Each problem is governed by a set of transition rules which can be stacked up to depth 3, and they find that GPT-3 performance largely surpasses human performance.

Many researchers also reject these claims, arguing that these examples do not accurately assess reasoning~\cite{mitchell_gpt3}, that LLMs cannot even solve the letter-string analogies attacked by the Copycat computer program 30 years ago~\cite{copycat_gpt3, copycat}, and that they still have ``a poor grasp of reality''~\cite{garymarcus}.

Given this debate, it is important to evaluate these models and whether they can already be used to solve ARC, which presents a formidable challenge compared to existing reasoning tasks such as in \cite{webb_emergent_2022}. To do this, we conduct experiments on several state-of-the-art LLMs by a new scheme to translate ARC tasks into a textual domain. We then evaluated them in the same framework as our DreamCoder solution and existing work. %

\subsubsection{Experimental Setup}

We evaluate several large language models on a common testing framework, designed to produce a fair comparison with existing ARC solutions. We test the OpenAI GPT series of models through their API, as well as Meta's LLaMA model series. For LLaMA models, we use GPTQ~\cite{gptq, gptq_github} to quantise the models to 4-bit precision, using a group size of 128. This can slight degrade performance on standard benchmarks but allows even the largest models to be executed on an NVIDIA A100 GPU.

\begin{figure}[h]
\centering
\includegraphics[width=\linewidth]{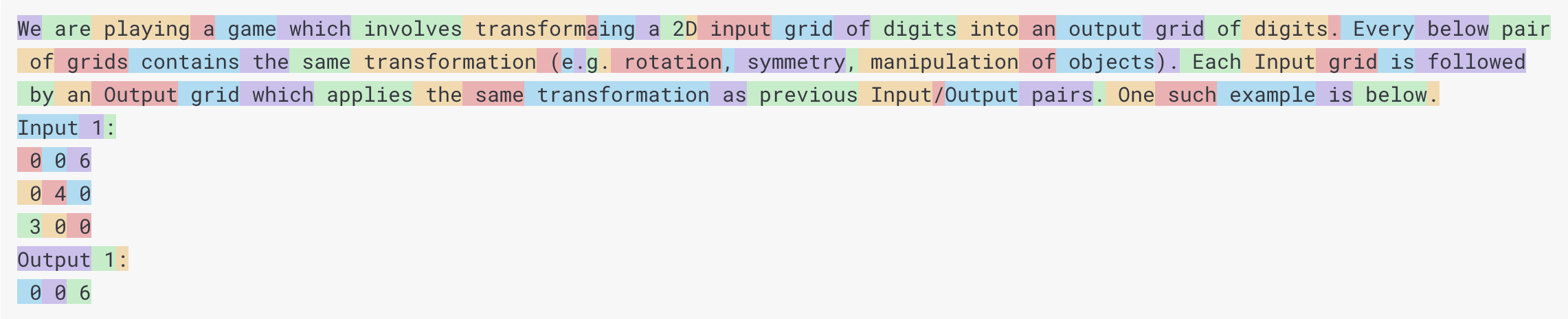}
\caption{An example of how the LLM `sees' an ARC task, with tokenisation highlighted (in this case using OpenAI Completion tokenisation). The task is structured so that each grid cell is a single token, with the test output left as a text completion.}
\label{fig:tokenisation}
\end{figure}

To encode ARC tasks in a textual format, we convert each grid into integer digits (representing colours), with newlines delimiting rows in the grid. The format is selected to match the tokenisation used by each LLM~\cite{openai-tokenizer}, which means that each grid cell corresponds to one token in the model. This is shown in \Cref{fig:tokenisation}. The LLM is fed a prompt explaining the problem, then the inputs and outputs for the training tasks, and the final grid representing the test task must be completed. Some LLMs are fine-tuned for a chat interface rather than text completion; for these models, we represent the input and output grids as a series of user/assistant messages with an overall system message describing the problem. Note that only problems that fit in 2048 tokens (the context window for most models tested) were attempted; in practice, we find that the largest problems are very difficult for LLMs and so this is likely to only marginally decrease performance. Full details of tokenisation and prompting are available in Supplementary Materials.

To combat the disadvantage that 1D sequence models would necessarily have on a 2D task format, we \textit{augment} each task with a transposed version and a 90-degree rotated version. This allows the LLM to attempt each task in a row-major and column-major format, dramatically improving overall performance.

\subsection{Understanding error cases in ARC}
\label{sec:error_cases}

Existing approaches to ARC broadly rely on the principle of program induction: generally, a DSL defines the space of possible solutions that an algorithm could ever produce, while some search process tries to induce programs. While DreamCoders model these two components explicitly, the principle applies to almost all current attempts to solve ARC, and similar reasoning algorithms such as Copycat~\cite{copycat}.

Furthermore, when a person attempts an ARC task, we see a similar process: one tries to abstract the training tasks to a `program' (generally in natural language in one's head, for example ``rotate by 90 degrees"); human intuition is the search procedure.

To guide our search for improved algorithms, we design a framework for analysing the shortcomings of existing algorithms. In the context of a program induction algorithm, we propose three classes of failure cases:

\begin{itemize}
    \item \textbf{Class 1:} The algorithm did not find a solution \textit{because the solution was not in the search space}. To resolve this, we need to increase the power of the algorithm by expanding the types of operations it knows (or can construct).
    
    \item \textbf{Class 2:} The algorithm did not find a solution, \textit{even though it was in the search space}. A prolonged search could have found the correct answer, but computational constraints meant it did not. In this case, we could increase available computation or find a way to guide the search towards promising avenues.
    
    \item \textbf{Class 3:} The algorithm found a candidate solution, \textit{but it did not generalize}. This is characterised by the finding of some rules that solve the training examples but yield an incorrect answer on the test example - essentially a false positive. This is perhaps the hardest case, as it cannot be solved by making the algorithm more complete - instead, we need to either reduce the search or design some notion of how plausible a given solution is so that they can be ranked. 
\end{itemize}

As an extreme example, we can consider an algorithm which enumerates random Python code to solve ARC tasks. This approach can never have a Class 1 error, because all ARC tasks can be solved in Python. However, it is likely to have many Class 3 errors (e.g., generating programs which simply return the training answers). In practice, we would encounter Class 2 errors because current computers could not hope to enumerate enough programs to find a correct one `by chance'.

Thus, while in principle a search algorithm sufficiently powerful to solve ARC tasks is easy to design (much like making a Turing-complete programming language is easy) - \textit{the difficulty is how to effectively prune and direct a search} such that it ``searches in the right direction" and can solve tasks in a reasonable time. This perhaps approximates something close to human intuition; when presented with an ARC task, a human does not mechanically attempt all possible transformations.

\subsection{Software}
\label{sec:software}

Throughout this work, the ability to quickly analyse and iterate through ARC tasks is instrumental. To this end, we have built the \code{arckit} python library, which contains tools to load and manage ARC and evaluate different algorithms. The library comes bundled with the ARC dataset, which can be rapidly loaded in a single line:

\begin{minted}{python}
easy_set, hard_set = arckit.load_data()
\end{minted}

Each dataset includes 400 \code{Task}s in a \code{TaskSet}, allowing them to be looked up by either index or ID. The \code{Task} class implements a number of features, such as scoring solutions and generating prompts for LLMs.

A primary strength of ARCkit is fast visualisation to understand the dataset and where algorithms fail; ARCkit can automatically produce vector graphics showing grids or tasks, exported in a PDF or SVG format, arranging grids to fit a specified figure size. This functionality was heavily used to generate the figures in this work. Furthermore, any task can be visualised in an interface using the \code{arctask} command or within Python. %

\enlargethispage{1\baselineskip}
The ARCkit library can be installed with the command \code{pip install arckit}, and documentation is available at \url{https://github.com/mxbi/arckit}. Additionally, we make our code for our DreamCoder implementation available from the following repository: \url{https://github.com/mxbi/dreamcoder-arc}.

\section{Results}

\subsection{Primitive design}
\label{sec:primitives}

The design of primitives for PeARL (\Cref{sec:methods:pearl}) %
has a big effect on the system's performance and is perhaps the most critical implementation detail. If primitives are too specialised or insufficiently powerful, tasks can be unsolvable. Conversely, if primitives are too elemental or there are too many non-useful primitives, the solutions may be too complex to find by search. Our design must balance these aspects for optimal performance (balancing Type 1 and Type 2 errors as described in \Cref{sec:error_cases}).

In existing applications of DreamCoder, the authors used relatively elemental primitives, highlighting the ability of abstraction sleep to discover higher-order behaviour~\cite{ellis_dreamcoder_2020}. For example, given list processing tasks (such as sorting a list or getting the maximum element), basic functional constructs such as \code{map, fold, cons, >} were defined, from which DreamCoder was able to learn essential constructs like \code{maximum} and eventually \code{sort}.

Given the difficulty of ARC tasks, which could limit abstraction sleep's ability to discover operations, a wide range or primitives were implemented in PeARL. Our domain-specific language (DSL) stems from an analysis of the existing state-of-the-art DSL, as well as an effort to categorise ARC tasks in a taxonomy (see Supplementary Figure \ref{fig:taxonomy}). This provides a good starting point for `powerful' primitives, which are likely to encode core knowledge priors (\Cref{sec:core_knowledge}).%
In addition, PeARL provides some basic primitives such as list processing, compositions and rotations, which allow DreamCoder to build new concepts during abstraction sleep.

\textbf{PeARL has 77 unique primitives}; below, we detail some broad categories. The full list is given in Supplementary Materials.

\paragraph{Rigid transformation \& Cropping}
Arbitrary rotations, flips and transpositions are all available, as well as cropping and uncropping operations. Grids can be sliced along an axis (for example, \ocaml{right_half}), and can be repeated or mirrored to solve symmetry tasks (these primitives are inspired by Alford et al. ~\cite{alford_thesis}).

\paragraph{Composition}
Grids can be stacked in an arbitrary order, optionally considering original positions. Pixelwise operations can operate on pairs of grids (e.g., \ocaml{pixelwise_and}).

\paragraph{Object manipulation}
A grid can be split into a \ocaml{list[grid]} in 5 distinct ways: 4/8-connectedness and based on rows, columns and colours. PeARL can apply an arbitrary function to each object in an image with the higher-order function \ocaml{mapSplit8}. Gravity can be simulated in a `Tetris-like' manner, and lines can be drawn between objects.

Inspired by Icecuber, a single object can be selected based on many attributes: size, frequency, density, colour and position. Lists of objects can be composed in order of size (\code{composeGrowing}), and PeARL can define and extend lists functionally (\ocaml{mklist, lcons}).

\paragraph{Colour manipulation}
PeARL can erase, filter and remap colours: both targeting a specific colour (\ocaml{c{1-9}}), as well as dynamically (\ocaml{{top,rarest}colour}).

\paragraph{Morphology}
PeARL can draw borders around objects, fill holes inside objects, compress blank spaces and repeated rows/columns.

\paragraph{Counting}
PeARL can count colours, pixels or objects in a grid (\ocaml{count{Pixels,Colours,Components}}), and use counts to construct new grids of a specified size (\ocaml{countTo{X,Y,XY}}).

\subsection{Evaluating DreamCoder}
\label{sec:dc_results}

To evaluate our DreamCoder solution, we perform two experiments, one on ARC-Easy and one on ARC-Hard. We run DreamCoder for a single wake-sleep cycle, training the recognition model for 40 minutes and attempting each task for 1 CPU-hour.%

Overall, DreamCoder enumerated 3.1 billion programs for 800 tasks, discovering frontiers for 75 and 23 tasks in ARC-Easy and ARC-Hard respectively. Of these frontiers, \textit{70 and 18 tasks were solved respectively}, with a Class 3 error rate (false positive solutions -- see \Cref{sec:error_cases}) of 1.25\%. These results dramatically improve on those achieved to date by DreamCoder (23 and 4 tasks solved by Alford et al. ~\cite{alford_thesis}). %

Overall, our implementation of DreamCoder achieves 16.5\% accuracy on ARC-Easy and 4.5\% on ARC-Hard, improving on other recent systems such as ARGA~\cite{arga}. However, a large gap remains between DreamCoder and the best hand-crafted solutions such as Icecuber. %

\begin{figure}[H]
\centering
\includegraphics[width=\linewidth]{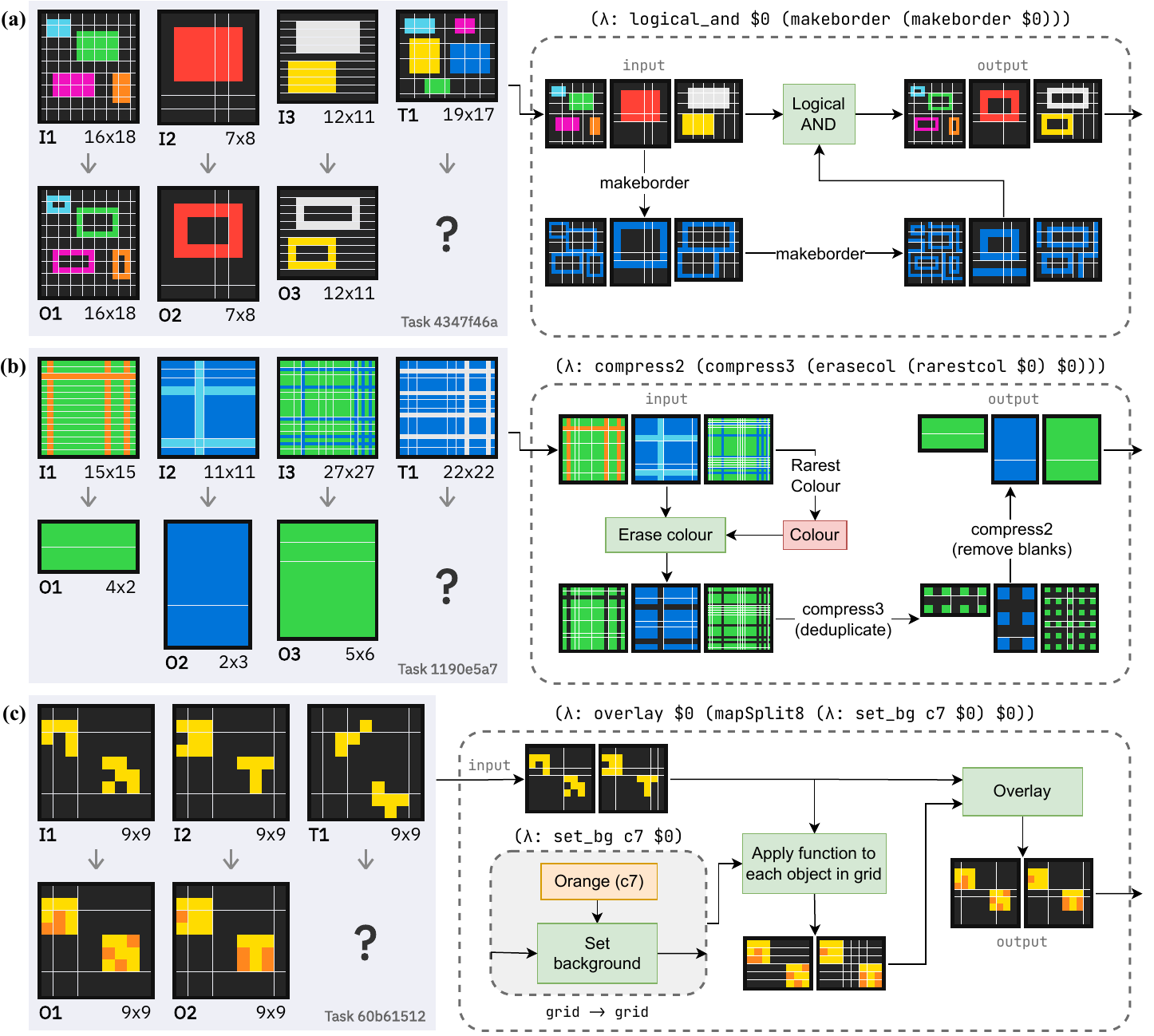}
\caption{Three ARC tasks solved by DreamCoder (left), with the PeARL programs it generated to solve them illustrated as a computation graph (right), with intermediate outputs shown. \\ \textbf{(a)} DreamCoder has no primitive to get the perimeter of objects, so it draws a border around it twice and takes the intersection with the original image. \textbf{(b)} Our type-system allows DreamCoder to operate on colours within images. \textbf{(c)} DreamCoder can use higher-order functions to apply arbitrary operations to each object in the image.}
\label{fig:dc_preds}
\end{figure}

\Cref{fig:dc_preds} shows three examples of PeARL solutions written by DreamCoder. DreamCoder can effectively use higher-order functions, types and combine primitives in interesting ways to discover new functionality. In the next section, we look at whether this performance would be achievable with brute force DSL search over PeARL, or whether dreaming and abstraction sleep bring improvements.

\subsection{Ablation Study}

The DreamCoder algorithm relies on both a good \textit{static} DSL and dynamic learning through abstraction and dreaming sleep. We conducted an ablation study to understand the impact of each learning phase. %

We ran three experiments: (i) with the recognition model disabled, (ii) with a simpler context-free recognition model that only models the probability of each primitive, and (iii) with the full contextual recognition model. The context-free model predicts the probability of individual primitives $f(\cdot)$ appearing in the solution, while the contextual recognition model also predicts the probability of every primitive within every context such as $g(\cdot, f(\cdot))$ We allowed 2 CPU-minutes of enumeration per task, and 10 minutes of recognition model training. The first DreamCoder iteration shows the algorithm's behaviour \textit{before} the abstraction sleep runs, with subsequent iterations incorporating compression.

\begin{figure}[h]
    \vspace{-2mm}
    \centering
    \captionsetup{justification=centering}
    \includegraphics[width=0.5\linewidth]{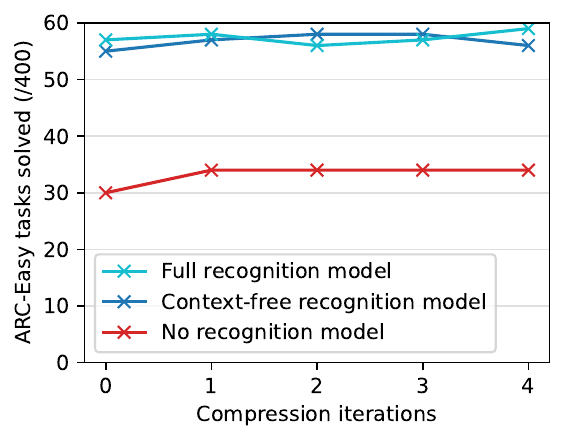}
    \caption{The number of ARC-Easy tasks solved by our DreamCoder implementation while ablating two key components. The addition of a neural network recognition model (abstraction sleep) doubles the number of solved tasks, while using compression to create new primitives by composing existing ones (dreaming sleep) has a much smaller performance uplift.}
    \label{fig:ablation}
\end{figure}

\Cref{fig:ablation} shows the effect of enabling the compression and/or recognition engine on ARC-Easy tasks solved. Compression (abstraction sleep) allows the search to solve a few more tasks by adding more powerful primitives, but the effect is smaller than expected. One explanation is that, unlike the original DreamCoder problems, PeARL was not intentionally designed to include only elementary primitives that compression could build upon, limiting the gain that composition can bring.

On the other hand, enabling the recognition model dramatically improved results, almost doubling the number of tasks solvable within the same time. \textit{This demonstrates the benefit that neural networks can bring to this problem.} We see that the full and context-free models perform similarly; it may be difficult to predict compositions of primitives directly from the input grids.

\enlargethispage{1\baselineskip}
Taking a closer look at the behaviour of abstraction sleep, we can identify why performance sometimes decreases in later iterations. The compression engine emits new primitives combining existing functions: \textit{these new primitives are assigned a lower entropy due to their frequent use in solutions}, and thus end up used in place of simpler counterparts that would solve the same solution - the result is that \textit{solutions to the same tasks tend to get longer} and thus more complicated over time, with more Class 3 errors. This suggests that the Minimum Description Length (MDL) principle used for ranking solutions could be modified to improve performance.

\subsection{Large language model results}

\begin{figure}[H]
\vspace{-5mm}
\includegraphics[width=\linewidth]{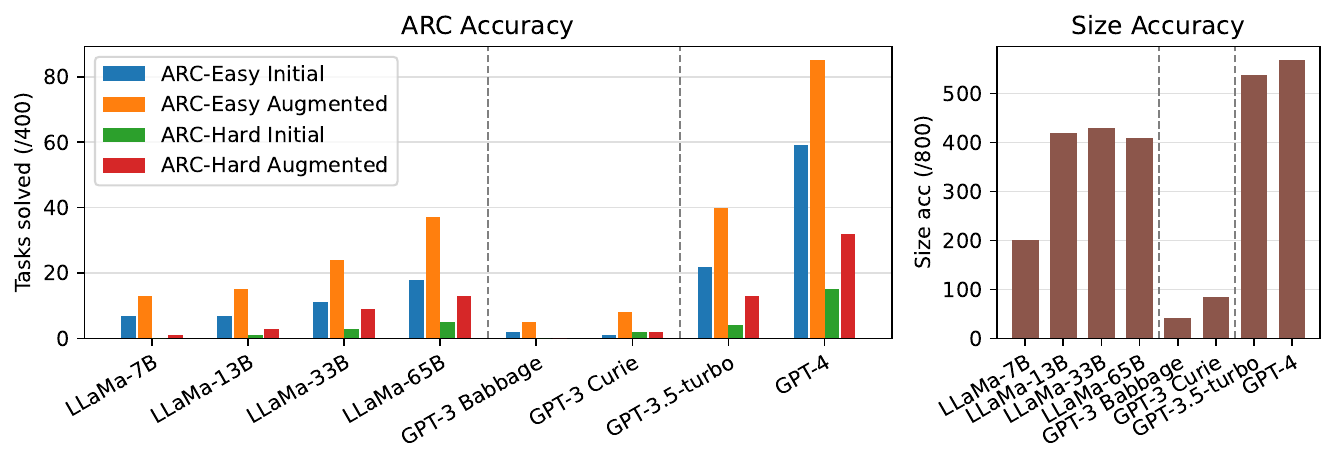}
\vspace{-8mm}
\caption{ARC performance of each large language model (LLM), as in Table \ref{tab:llm_results}. The number of correctly solved tasks is given for ARC-Easy and ARC-Hard, both on the original task and our 3-shot augmentation scheme. Size Accuracy gives the number of tasks for which the LLM was able to output any grid of the correct size across both datasets.}
\label{fig:llm_results}
\end{figure}

We test a variety of large language models (LLMs) on ARC. Table \ref{tab:llm_results} shows the results on ARC for some popular LLMs. Figure \ref{fig:llm_results} shows the accuracy attained by each LLM on each ARC dataset, broken down into three accuracies: \textbf{initial} shows the proportion of tasks which are solved on a first attempt; \textbf{augmented} allows three attempts with rotated and transposed tasks following our augmentation scheme. Finally, we consider a `size accuracy' which shows whether the LLM outputs a grid of the correct size (even if the contents are incorrect).

\footnotetext[1]{The size of these models are a trade secret, but GPT-4 likely exceeds 175B parameters.}

\begin{table}[h]
\centering
\caption{Results on ARC for 8 popular LLMs. {\bf Init.} gives the initial 1-shot accuracy, {\bf Aug.} allows the model three guesses using our augmentation scheme. {\normalfont Size acc} gives the proportion of tasks in which a correct-size grid was predicted across both datasets.}
                        
\renewcommand{\solved}[1]{#1}
\begin{tabular}{@{}clrrrrrr@{}}
\toprule
& & & \multicolumn{2}{c}{ARC-Easy/400} & \multicolumn{2}{c}{ARC-Hard/400} & Size acc \\ \cmidrule(lr){4-5} \cmidrule(lr){6-7}
Series            & Model        & Params  & \textbf{Init.} & \textbf{Aug.} & \textbf{Init.} & \textbf{Aug.} & /800 \\ 
\midrule
\multirow{4}{*}{Meta}   & LLaMA-7B~\cite{llama} & 6.7 B & \solved{7} & \solved{13} & \solved{0} & \solved{1} & \solved{201} \\
                        & LLaMA-13B~\cite{llama} & 13 B & \solved{7} & \solved{15} & \solved{1} & \solved{3} & \solved{334}       \\
                        & LLaMA-33B~\cite{llama} & 33 B & \solved{11} & \solved{24} & \solved{3} & \solved{9} & \solved{429} \\
                        & LLaMA-65B~\cite{llama} & 65 B & \solved{18} & \solved{37} & \solved{5} & \solved{13} & \solved{408} \\ \midrule
                        
\multirow{2}{*}{\shortstack{OpenAI\\ Complete}} & GPT-3 Babbage~\cite{gpt3} & 1.3 B & \solved{2} & \solved{5} & \solved{0} & \solved{0} & \solved{42}       \\
                        & GPT-3 Curie~\cite{gpt3}   & 6.7 B & \solved{1} & \solved{8} & \solved{2} & \solved{2} & \solved{85}     \\ \midrule
\multirow{2}{*}{\shortstack{OpenAI\\ Chat}}  & GPT-3.5-turbo~\cite{chatgpt} & -\footnotemark[1]     & \solved{22} & \solved{40} & \solved{4} & \solved{13} & \solved{538}      \\
                        & GPT-4~\cite{gpt4}          & -\footnotemark[1]     & \solved{59} & \solved{85} & \solved{15} & \solved{32} & \solved{567}     \\ \bottomrule
\end{tabular}

\label{tab:llm_results}
\end{table}

Several broad conclusions can be drawn from these results. First, we see a prominent performance scaling as the model size increases, especially in the LLaMA models, where progressively larger models are trained similarly. It is likely that this trend continues past currently available models, but the extent to which this continues is unclear and an open research question~\cite{chinchilla_scaling}. We also see that the ARC-Hard dataset is much more challenging, with small models solving almost no tasks. 

Overall, the GPT-4 model from OpenAI massively outperforms all other LLMs, solving 21\% of easy tasks and 8\% of hard tasks. We see that across all models, \textbf{our augmentation can double accuracy or more} (Figure \ref{fig:llm_results}). The smaller GPT-3 models severely underperform, while large LLaMA models have middling performance. 

\begin{figure}[H]
\centering
\includegraphics[width=\linewidth]{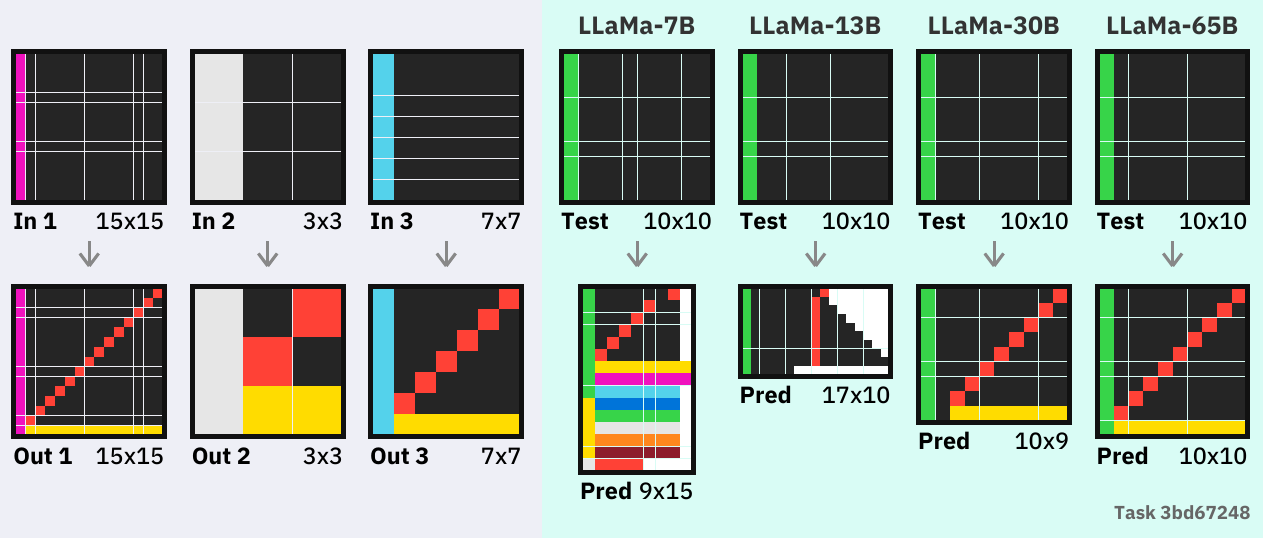}
\caption{An example task solved by the large-language model LLaMA-65B but not by smaller models; training examples are on the left with predicted answers on the right. We see the smallest two models struggle to produce a coherent answer. The 30B model is almost correct (skipping a row and one yellow pixel), and the largest model solves the task perfectly.}
\label{fig:llm_example1}
\end{figure}

\begin{figure}[H]
\centering
\begin{subfigure}{0.519\linewidth}
\includegraphics[width=\linewidth]{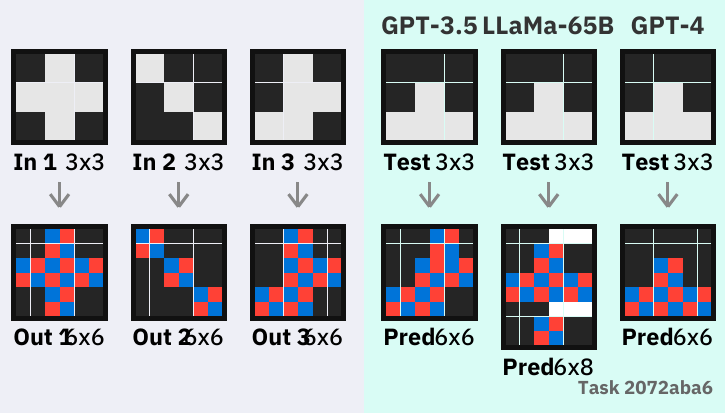}
\end{subfigure}
\begin{subfigure}{0.471\linewidth}
\includegraphics[width=\linewidth]{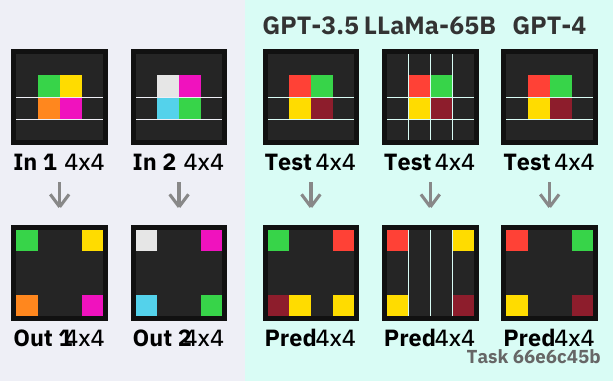}
\end{subfigure}
\caption{Two ARC-Hard tasks where only GPT-4 produced a correct solution. While the other LLMs often produce somewhat close matches on first inspection, an exactly correct solution is required to solve a task.}
\label{fig:llm_example2}
\end{figure}

Finally, we see that the size accuracy gives an easier metric that correlates with overall performance: models that solve very few tasks can be compared by their size accuracy. \Cref{fig:llm_example1,fig:llm_example2} qualitatively show the types of errors that underperforming LLMs make. %

Overall, we see that with our encoding and augmentation scheme, GPT-4 achieves slightly better accuracy than our DreamCoder solution (see \cref{sec:dc_results}). %

\subsubsection{Applicability of LLMs to ARC problems}

One fair criticism of evaluating LLM abilities on ARC is that these models are designed to be text-based; they read a stream of tokens. However, we argue that this still provides a useful comparison point: there is no known equivalent to LLMs that could support reasoning with visual input and output data. This is because LLMs derive their abilities from the huge diversity of tasks displayed in their training data (often encompassing a trillion tokens of code, scientific papers, forums, etc.), and performance has been shown to be very dependent on training data~\cite{llama}. In the visual domain, equivalent datasets that display broad human reasoning capabilities are much harder to collect, although some promising work on Large Vision Models has shown abilities on other reasoning tasks~\cite{lvm}. There are also several alternative ways of using LLMs that could be investigated. We discuss some of these in \Cref{sec:future_work}.

\subsection{Ensemble methods}
\label{sec:ensemble}

In this work, we evaluate some very different systems as ARC solvers. One interesting question arises when looking at the differences between these systems: are they solving the same tasks, or does each system have different strengths?

Machine learning research has looked at ensemble approaches, where multiple models are combined to produce predictions better than any single model. The success of these approaches relies on \textit{diversity} of models~\cite{diversity_ensemble1}, often measured by looking at the correlation of predictions. More diverse models produce more powerful ensembles, even with disparate individual accuracies. %

Since we cannot measure correlation between predictions directly, we propose two approaches to measure the similarity between systems attempting ARC, even when models have very different headline performance. First, the overlap between the set of tasks solved by each system (the Szymkiewicz–Simpson coefficient~\cite{ss_coef})

$$\mathrm{Overlap}(A, B) = \frac{|A \cap B|}{\min(|A|, |B|)}$$

tells us the proportion of tasks in the weaker model solved by the stronger model. When one model solves a subset of the tasks of another, the overlap is 1. This can be seen as a performance-normalised similarity metric: it ignores the overall accuracy difference between the models, and lets us determine if one system is a more powerful version of another, or whether they are solving complementary types of tasks (as opposed to more common metrics such as Jaccard index). We also propose the asymmetric Gain measure,

$$\mathrm{Gain}(A, B) = |A \cup B| - |A|,$$

which gives the additional tasks solvable by adding one model to another model - hence, it reflects the \textit{unique skills} of each model; tasks that cannot be solved by the other model in the pair. The asymmetry means that we can see which model performs the best.

\begin{figure}[H]
\vspace{-3mm}
\begin{subfigure}[t]{0.48\linewidth}
    \vskip -4pt
    \includegraphics[width=\linewidth]{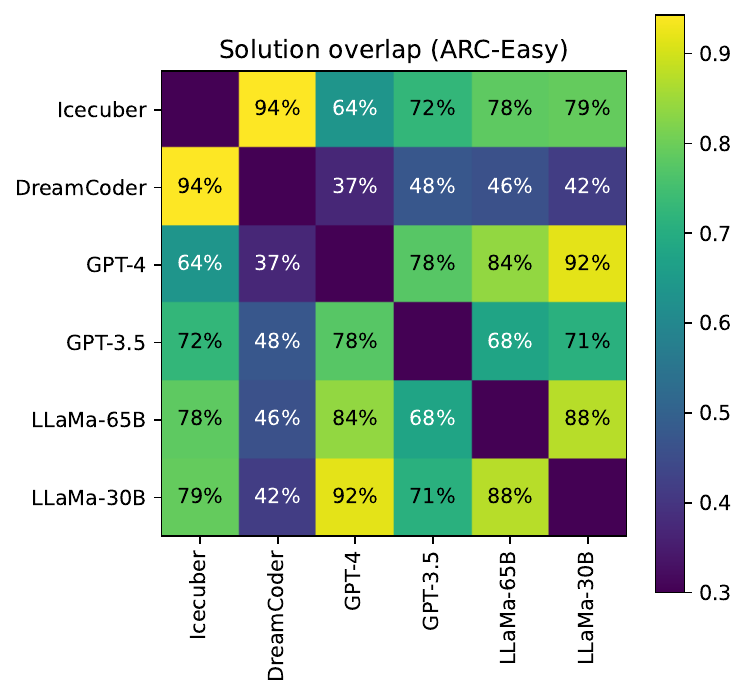}
\end{subfigure}
\begin{subfigure}[t]{0.50\linewidth}
    \vskip 0pt
    \includegraphics[width=\linewidth]{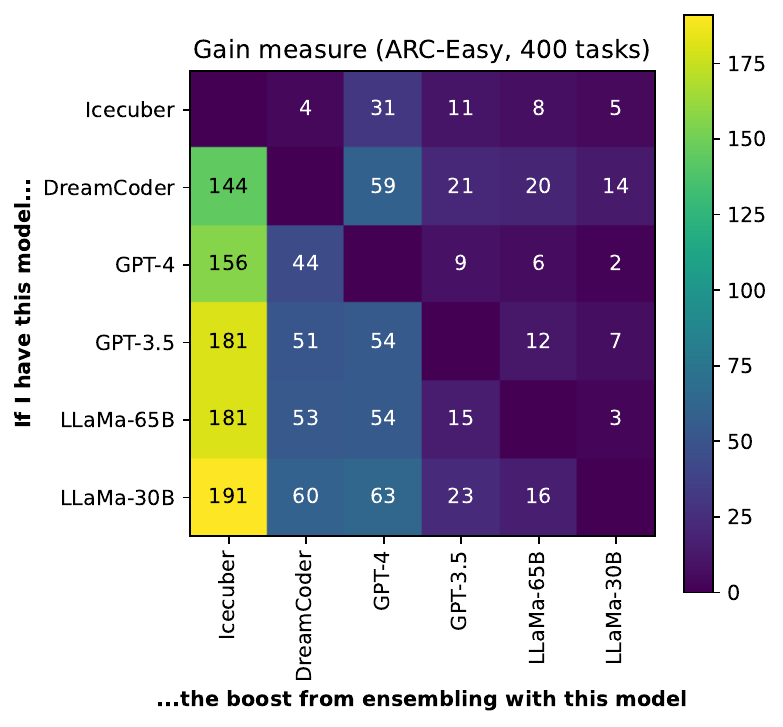}
\end{subfigure}
\caption{Comparison of the solution similarity of our systems, showing solution overlap (left panel) and gain measure (right panel) between each system: this tells us how similar the predictions are, and how much one could expect to gain by combining systems. Systems are ordered by overall performance, so the top-right triangle in the panel on the right shows the unique skills of the lesser system. The panel on the left shows that DreamCoder and GPT-4 have only 37\% overlap between the tasks which they can solve.}
\label{fig:ensemble}
\end{figure}

Figure \ref{fig:ensemble} shows both metrics for our systems' predictions on the ARC-Easy set, from which we can make a few key takeaways. %

First, we see that LLMs tend to solve different tasks than DSL-based solutions: the overlap between these models is relatively low. In particular, our DreamCoder and GPT-4 solutions, which have comparable accuracy independently, have only a 37\% overlap between \textit{which} tasks they can solve - this suggests an ensemble which can effectively select between approaches \textit{could lead to an almost doubling in performance}.

Conversely, we see that different LLMs tend to have similar predictions (with larger models tending to solve a superset of problems solved by their smaller counterparts). Likewise, we see that DreamCoder and Icecuber solve similar types of task: with DreamCoder uniquely solving only 4 tasks and a 94\% overlap between them.

\subsection{Building a heterogenous high-performance ensemble}

For classification and regression tasks, ensembles can be built effectively by a weighted average of each system's predictions~\cite{diversity_ensemble1}. In our case, with only exact solutions scoring credit, interpolating solutions would be unlikely to work well.

Instead, we build a \textit{voting} ensemble of systems: on each task, each system can propose some output grids which they `vote' for, added to a priority queue. When multiple systems generate the same prediction, votes are summed, increasing their prority. Since ARC allows three guesses, we select the three output grids with the most votes. Another difficulty to overcome is the heterogeneity of our systems. For example, DreamCoder only produces a prediction when it solves all training tasks: a DreamCoder prediction should be prioritised for the final list. On the other hand, when combining LLMs, we would like to provide higher weight to LLMs with better performance.

We assign distinct weights to each algorithm to build our ensembles: 20 to DreamCoder, 3 to each GPT-4 prediction, and $[8, 4, 4]$ to the three IceCuber predictions for each task (based on inspection of the above figures and minimal manual tuning). Additionally, we discard any LLM predictions which are not grids, the default prediction from Icecuber (predicted when no solution was found), and de-duplicate predictions from DreamCoder (as it tends to find equivalent programs: e.g., when functions are commutative under composition, we have $f(g(x))$ and $g(f(x))$). This process allows us to select the three most promising predictions from up to 9. The performance of our ensembles are discussed in the next section.

\pagebreak
\subsection{Quantitative results across all methods}
\label{sec:quantitative}

\begin{figure}[h]
\centering
\includegraphics[width=\linewidth]{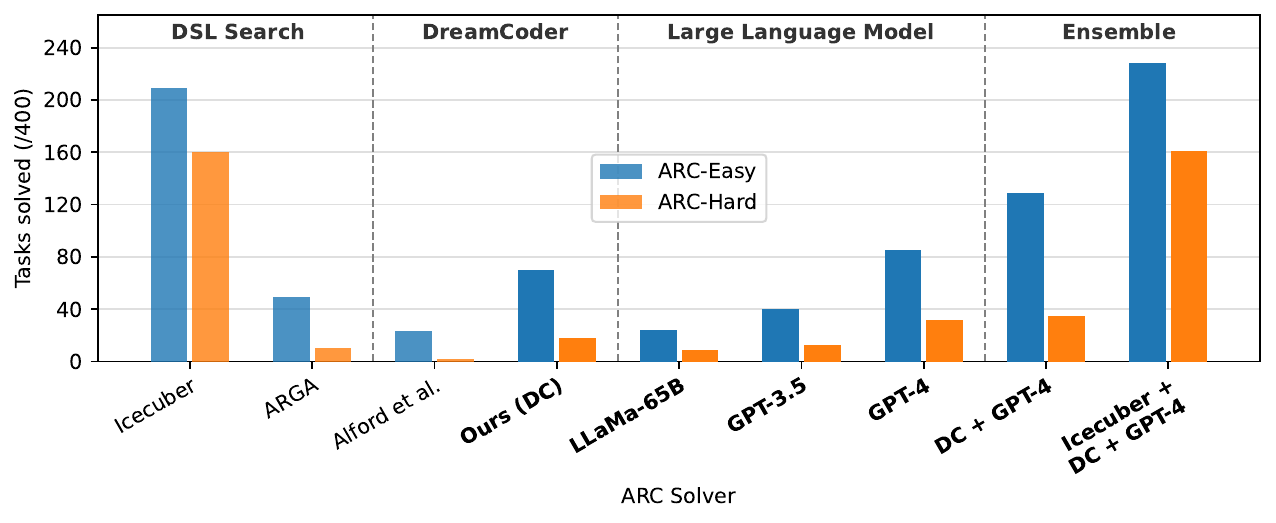}
\vspace{-8mm}
\caption{ARC tasks solved by each system in this work on the two available ARC datasets (see \Cref{tab:results}); results from this work in bold. The fully handcrafted Icecuber solver remains the best single solution, with GPT-4 obtaining slightly better performance than DreamCoder. We achieve dramatic gains by ensembling solutions. For the sake of brevity, DreamCoder is abbreviated to DC and Abstract Reasoning with Graph Abstractions~\cite{arga} is abbreviated to ARGA.}
\label{fig:results}
\end{figure}

We now take stock of each system discussed in this report, looking at headline performance on both ARC datasets. For fairness, all systems were evaluated from scratch using a common framework, following the official ARC evaluation procedure. Each system has access to training examples, and can create a maximum of three predictions for each test example. A task is solved if one of the three attempts are \textbf{exactly} correct. \Cref{fig:results} and \Cref{tab:results} show these results.

\newcommand{\ts}[1]{}

\renewcommand{\solved}[1]{#1}
\begin{table}[h]
\centering
\caption{Comparison of headline performance for the systems discussed in this work; all systems have three attempts per task. Systems in \textbf{bold} were designed in this project. For the sake of brevity, DreamCoder is abbreviated to DC and Abstract Reasoning with Graph Abstractions ~\cite{arga} is abbreviated to ARGA.}
\begin{tabular}{rcrr}
\toprule
Type & System & \hspace{-1cm}ARC-Easy/400 & ARC-Hard/400 \\\midrule
\multirow{2}{*}{DSL Search (\cref{sec:related_work})} & Icecuber~\cite{icecuber_blog} & \solved{209} & \solved{160} \\
 & ARGA~\cite{arga} & \solved{49} & \solved{10} \\

\midrule\multirow{2}{*}{DreamCoder (\cref{sec:dreamcoder})} & Alford et al.~\cite{alford_thesis} & \solved{23} & \solved{2} \\
 & \textbf{Ours (DC)} & \solved{70} & \solved{18} \\
 
\midrule\multirow{3}{*}{LLM (\cref{sec:llms})} & \textbf{LLaMA-65B}~\cite{llama} & \solved{24} & \solved{9} \\
 & \textbf{GPT-3.5}~\cite{chatgpt} & \solved{40} & \solved{13} \\
 & \textbf{GPT-4}~\cite{gpt4} & \solved{85} & \solved{32} \\

\midrule\multirow{2}{*}{Ensemble (\cref{sec:ensemble})} & \textbf{DC + GPT-4} & \solved{129} & \solved{35} \\
 & Icecuber + \textbf{DC + GPT-4} & \textbf{\solved{228}} & \textbf{\solved{161}} \\\bottomrule
\end{tabular}
\label{tab:results}
\end{table}

Our new approaches GPT-4 and DreamCoder attain decent performance, improving over ARGA~\cite{arga}. Notably, DreamCoder solution solves $3\times$ and $9 \times$ more ARC-Easy and ARC-Hard tasks respectively than the existing best implementation by Alford \etal ~\cite{alford_thesis}. %

Our voting ensemble can combine the strengths of different algorithms, with DreamCoder and GPT-4 ensembling to solve \textbf{50\% more tasks than either system alone}. Moreover, these systems successfully tackled tasks that stumped Icecuber, the previous state-of-the-art solution. Our final ensemble, which incorporates Icecuber, DreamCoder, and GPT-4, manages to solve 57\% and 40\% of tasks on the ARC-Easy and ARC-Hard datasets respectively.%

Across the board, performance on ARC-Easy correlates well with ARC-Hard, with most systems solving $2-4$ times more easy tasks - a clear outlier here is Icecuber, which solves 160 hard tasks. This could be partially attributed to the fact that \textbf{the Icecuber DSL was built on hand-crafted solutions to 100 hard tasks}~\cite{icecuber_blog}, meaning that the developer designed the language to be good on these tasks.

Collectively, these results show two promising approaches using neural networks for abstraction and reasoning. We highlight the need for diverse and complementary methods that, when combined, can lead to superior performance. While Icecuber, the current state-of-the-art that uses hand-crafted brute-force search retains a commanding lead over neural-network-based solutions, we significantly close the gap. 

\section{Discussion}
\label{sec:future_work}

\subsection{Overview}

In this work, we have applied a modern machine learning lens to a touchstone problem in AI: searching for systems that can perform \textit{broad generalization} to abstraction and reasoning tasks. Overall, we find that complex human-crafted solutions based on domain specific languages (DSL) search still provide the best known single-system performance. However, we demonstrate two promising machine-learning-based solutions, which provide a concrete research direction to make more robust learning machines.

\subsection{Limitations of our approach}

There are several types of ARC tasks which our DreamCoder adaptation cannot solve. For example, \textit{copy-paste} tasks, which rely on copying a pattern from training outputs to test outputs, cannot be solved. This is because DreamCoder considers each input/output pair in isolation, and cannot copy objects from one grid to another. \textit{In-painting} tasks can likely be solved much more effectively by a dedicated algorithm, which tries to identify the tiling or symmetry pattern required to fill in the missing pixels.

Our method for adapting large-language models to ARC is also exploratory, and there is likely to be significant gains attainable from the use of different models, different prompt techniques (for example that allow LLMs to solve problems larger than their context window), or systems that combine mechanistic reasoning with LLMs. We discuss some avenues in the next section.

\subsection{Future work}

Research on computational abstraction and reasoning is far from complete, so we consider some future directions suggested by our conclusions. 
\begin{itemize}
\item While we improve the accuracy of DreamCoder, it is still far behind the handcrafted Icecuber DSL search~\cite{icecuber_blog}. One future approach would be to extend DreamCoder with the unique features of Icecuber, such as colour normalisation and in particular the greedy stacker, which combines intermediate solutions working backwards from the output grid. Execution-guided program synthesis would be one way to achieve a similar effect, where intermediate solutions are scored based on their similarity to the final output and used to guide the search, replacing all-or-nothing evaluation.
\item There are several research avenues that could improve LLM performance. We only evaluate the in-context learning performance of LLMs; advances such as low-rank adaptation fine-tuning (LoRA) ~\cite{lora} could precondition a model further on ARC priors. %
Additionally, chain-of-thought prompting, where the LLM is first required to explain the training examples before applying it to the test example, may also improve performance ~\cite{chain_of_thought}. 

\item Early experiments have been made with Large Vision Models~\cite{lvm} which suggest that visual datasets are sufficient to train a model with emergent reasoning abilities that could be applied to ARC. While these models warrant investigation, challenges are likely to include the requirement for pixel-perfect outputs as well as collecting a suitable mix of training data. %
\item An ARC2 dataset is in development by Chollet and Lab42~\cite{arc2}, aiming to crowd-source 5,000 tasks. ARC2 aims to be more diverse, with ``no specific task type used more than once''. This is likely to make ARC even harder, and require greater generalization abilities.
\end{itemize}

\subsection{Concluding remarks}

Our implementation of DreamCoder achieves a \textbf{$3.5\times$ }improvement in tasks solved across ARC over the previous best implementation~\cite{alford_thesis, dreaming_arc}, through a wide range of improvements, such as the PeARL language and the use of a neural-network recognition model (\cref{sec:dreamcoder}).

Additionally, we introduced a framework to transform visual ARC problems into a text completion domain suitable for large language models (LLMs), and found that the largest available LLMs trained across huge corpora have sufficiently transferable abilities to begin to solve ARC problems; their accuracy can also be effectively doubled with suitable augmentation. Our results suggest that even larger models could continue to improve attainable accuracy on ARC (\cref{sec:llms}). While previous solutions largely rely on human-designed priors in the form of a DSL, the allure of LLMs is that they instead use neural-networks in an end-to-end fashion, and all reasoning priors come from training data. %

We also show the effectiveness of ensembling solutions: while each algorithm may specialise on a certain type of task, systems with low false-positive rates can be effectively combined (\cref{sec:ensemble}). Rather than trying to create a system that \textit{beats} Icecuber, a useful direction would be to try to build systems that complement the state-of-the-art, leading to a multipronged solution with better generalization. In particular, we already find that LLMs solve many problems not suitable for domain-specific languages. %

It is also sobering to observe that the previous state-of-the-art solutions as well as the approaches developed in this work still give us only modest accuracy on the ARC dataset (approximately 40\% tasks solved on the ARC-Hard dataset). This is still much less than what humans can solve~\cite{arc,Mitcell_arc_humans,human_performance_arc}.

Finally, we see that broad generalization is still an incredibly challenging goal for AI, even for the most advanced systems. ARC is proving to be an informative benchmark; still unsolved, and motivating research after four years, while at the same time well-defined and objectively evaluable. We hope that research on ARC will one day bring about a broad generalization machine, in the way that MNIST and ImageNet brought about the rise of modern deep neural networks~\cite{imagenet_transform}.%

\section*{Declarations}

\subsection*{Acknowledgements}
We acknowledge the help and support of the Accelerate team. We are especially grateful to Neil Lawrence, Carl Henrik Ek and Jessica Montgomery for fruitful discussions and feedback. %

\subsection*{Funding statement}
SB acknowledges funding from the Accelerate Programme for Scientific Discovery Research Fellowship. The funders had no role in study design, data collection and analysis, decision to publish, or preparation of the manuscript. The views expressed are those of the authors and not necessarily those of the funders.

\subsection*{Conflicts of interests}
All authors declare they have no conflicts of interest to disclose.

\subsection*{Ethics}
No ethics approval was necessary. %

\subsection*{Data accessibility}
This study does not generate any data.

\subsection*{Author contributions}

\enlargethispage{1\baselineskip}
MBI carried out the implementation and analysis, participated in the design of the study and wrote the manuscript. SB carried out the analysis, participated in the design of the study and wrote the manuscript. SB directed the study. All authors gave final approval for publication.%

\bibliography{refs}

\pagebreak

\pagebreak

\pagebreak

\newpage

\section*{Supplementary Material}

\subsection*{A taxonomy of tasks in ARC}
\label{sec:taxonomy}

The first step to constructing an ARC solver is to understand the nature of the problem: this helps motivate later design choices. To do this in a systematic way, we built a \textit{taxonomy} of ARC tasks. Some example tasks in this taxonomy are shown in Supplementary Figure \ref{fig:taxonomy}. Understanding common concepts is vital to designing a good domain-specific language (DSL) to solve tasks that involve these concepts (see \Cref{sec:primitives}).

\begin{suppfigure}[H]
\centering
\includegraphics[width=0.97\linewidth]{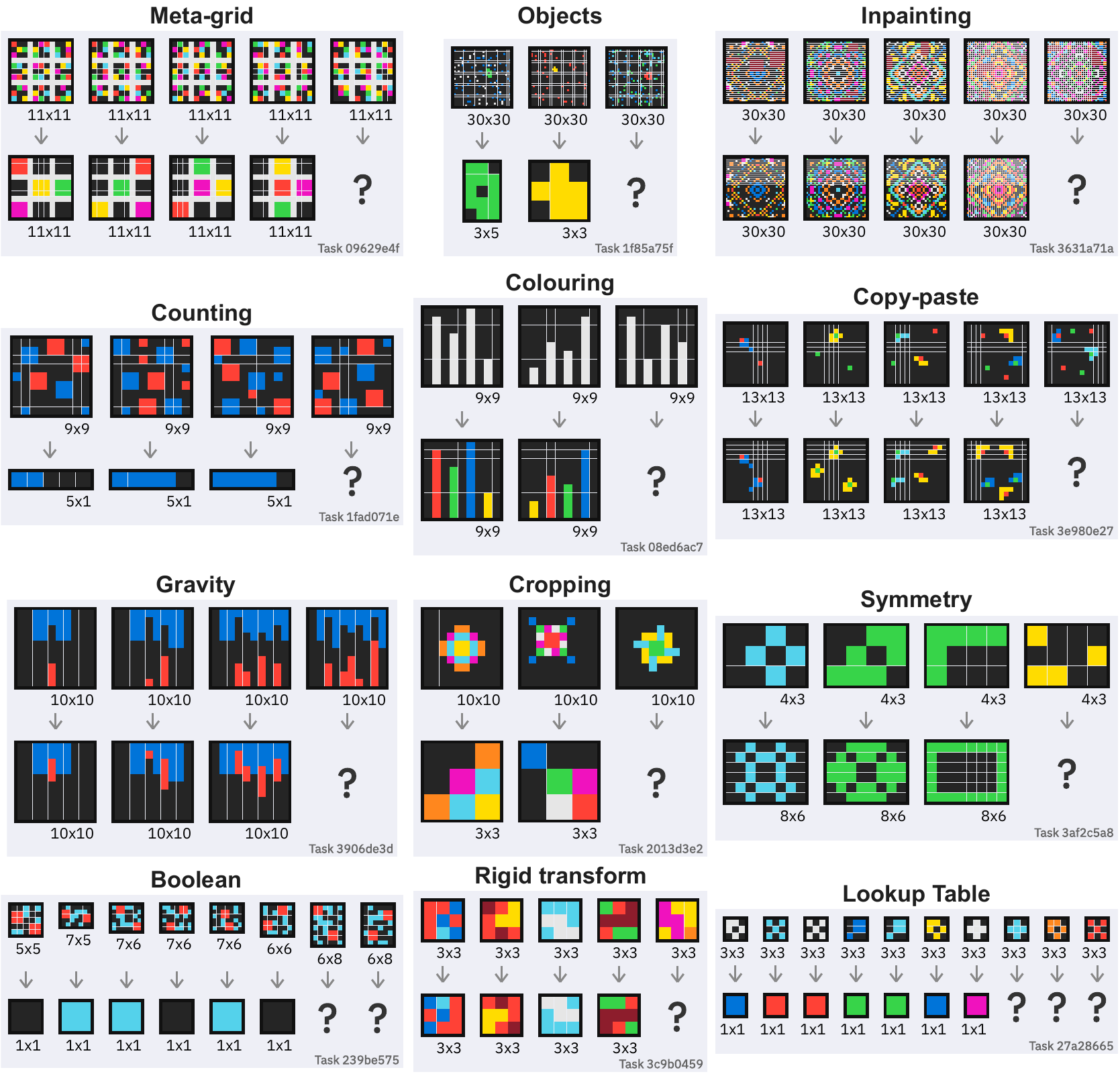}
\caption{Example tasks for 12 taxonomy tags: the diversity of ARC is apparent. Many tasks combine multiple of these concepts or other concepts not shown.}
\vspace{-3mm}
\label{fig:taxonomy}
\end{suppfigure}

\subsection*{Icecuber: DSL Search}

The existing state-of-the-art approach for ARC was introduced by Johan Sokrates Wind (a.k.a \textbf{Icecuber}), achieving a 20.6\% private test set accuracy~\cite{icecuber_blog} in the first Kaggle competition, and has yet to be surpassed.

Icecuber implements a Domain-Specific Language (DSL) in C++, with 42 image transformation functions with 142 total variants~\cite{icecuber_code} (for example, the \code{eraseColor} function has variants for 10 colours). Each function is a unary transformation from one grid to another, such as cropping, filling in the interior of objects, or re-colouring a grid. 

The DSL is combined with a highly efficient brute-force search written in C++, stacking up to 4 unary functions for each task (\Cref{fig:icecuber}). Rather than storing the programs that generate each specific outcome, each function is applied to the entire set of training and test inputs at once (the starting piece), creating a new piece stored in a directed acyclic graph (DAG). After all programs are enumerated, this DAG contains as many as $10^7$ pieces (candidate output grids for each input grid). The DAG allows for de-duplication of outputs, which helps with memory usage and performance (\eg \code{rotate180(rotate180(grid))} points back to the input piece in the DAG, and therefore is not enumerated any further).

Each piece can then be checked against the training data: if all the training inputs are transformed correctly, we have found a candidate solution to our task (and the corresponding test output is already in the piece).

When none of the candidates exactly match the training outputs, the \textbf{greedy stacker} can instead be employed. The greedy stacker works by choosing a random subset of training examples, and \textbf{selects pieces from the DAG that most closely match the unexplained cells in the grid}, composing them with transparency to minimise the Hamming distance. This allows bidirectional search: finding intermediate states that compose to the output. Finally, solutions are ranked by a complexity heuristic and the top three solutions are submitted.

Several additional tricks are employed, such as normalising colours in tasks, and a module which guesses the output grid size before enumeration begins.

By limiting the DSL to unary functions only, the search space size can be greatly limited, and becomes amenable to brute-force search. A more complicated DSL that can arbitrarily combine grids would suffer from super-exponential growth, dramatically reducing the maximum depth possible. However, the greedy stacking solver is able to overcome the drawbacks of this by combining grids \textit{after enumeration}, reducing the overall computational complexity.

\pagebreak
\subsection*{Prompting LLMs to complete ARC tasks}
\label{appendix:llms}

\subsubsection{Tokenisation}

For each type of LLM, we use a different grid-encoding scheme to match the tokeniser used by that LLM. 

\begin{description}
    \item[LLaMA] We encode grids with no spaces between digits, and with newlines between rows. The LLaMA tokeniser creates a separate token per numerical digit.
    \item[OpenAI Completion] These models use the GPT-2 tokeniser, which has unique tokens for each digit preceeded by a space, so we add a space before each cell to maintain one token per grid.
    \item[OpenAI Chat] These models use the CL100K tokeniser, which \textbf{has no unique tokens for digits followed/led by static characters}; therefore we do not have a way to force each cell to be a single token. We use a tokenisation with no spaces and therefore allow multiple digits to be merged into one token.
\end{description}%

\subsubsection{Prompting}

\paragraph{Completion prompt format}

We use the following prompt for completion models:

\begin{minted}[fontsize=\footnotesize, bgcolor=mintedbg, breaklines,breaksymbolleft=]{text}
We are playing a game which involves transforming a 2D input grid of digits into an output grid of digits. Every below pair of grids contains the same transformation (e.g. rotation, symmetry, manipulation of objects). Each Input grid is followed by an Output grid which applies the same transformation as previous Input/Output pairs. One such example is below.
\end{minted}

\vspace{-1cm}

\paragraph{Chat prompt format}

Chat models are more complex and use a \textit{system} message followed by interleaved user and assistant messages. We use the following system message:

\begin{minted}[fontsize=\footnotesize, bgcolor=mintedbg, breaklines,breaksymbolleft=]{text}
We are playing a game which involves transforming an input grid of digits into an output grid of digits. In general, digits form objects in 2D and the task is to perform some spatial transformation of these objects to go from the input grid to the output grid. All the information about the transformation is contained within the input pairs themselves, and your answer will only be correct if the output grid is exactly correct, so this is what I expect from you. I will begin by giving you several examples of input-output pairs. You will then be given a new input grid, and you must provide the corresponding output grid.
\end{minted}

\vspace{-5mm}

We then interleave the input and output grids as User/Assistant messages, so that the chat model can follow the behaviour of what are presented as its previous responses.

We always use greedy sampling when generating predictions to avoid 

\subsection*{DreamCoder Software Architecture}

\begin{suppfigure}[H]
    \label{fig:dreamcoder_overview}
    \centering
    \includegraphics[width=0.9\linewidth]{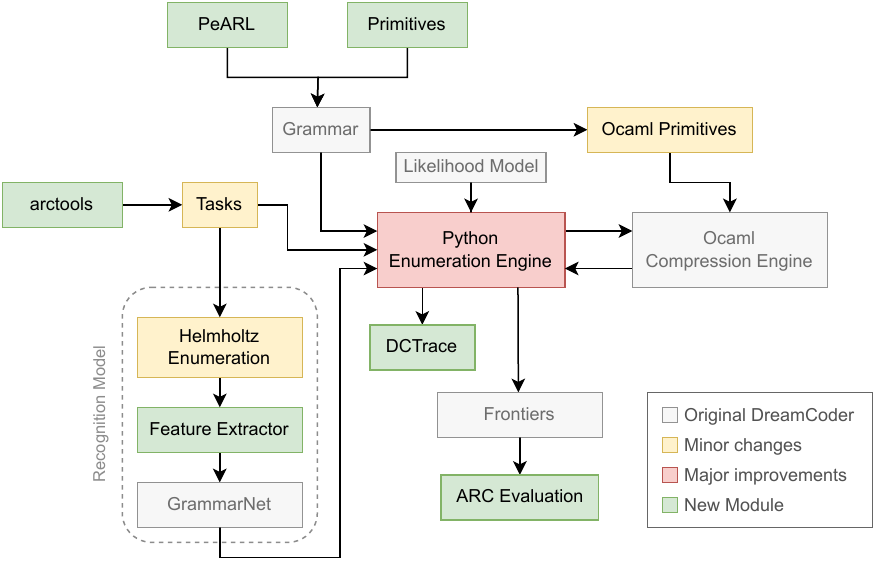}
    \caption{An overview of the DreamCoder software architecture, highlighting the primary modules used, and whether they are new or existing modules.}
    \label{fig:dreamcoder_overview}
\end{suppfigure}

\pagebreak
\subsection*{PeARL Primitives}
\label{appendix:dsl}

The following list provides a complete description of the primitives available in PeARL. A number of these were adapted from the Icecuber C++ DSL~\cite{icecuber_blog}.

\newcommand{\primname}[1]{\ocaml{#1}}
\newcommand{\primtypesig}[1]{\ocaml{#1}}

{\fontsize{7pt}{7pt}\selectfont
\begin{tabular}{lll}
\toprule
\textbf{Primitive} & \textbf{Type} & \textbf{Description}\\\midrule

\textbf{Rigid transformations}\\
\primname{rot90} & \primtypesig{grid -> grid} & Rotate 90° clockwise\\
\primname{rot180} & \primtypesig{grid -> grid} & Rotate 180°\\
\primname{rot270} & \primtypesig{grid -> grid} & Rotate 90° counter-clockwise\\
\primname{flipx} & \primtypesig{grid -> grid} & Horizontal flip\\
\primname{flipy} & \primtypesig{grid -> grid} & Vertical flip\\
\primname{swapxy} & \primtypesig{grid -> grid} & Transpose\\

\midrule
\textbf{Cropping}\\
\primname{left_half} & \primtypesig{grid -> grid} & Crop the left half of the grid (floor division)\\
\primname{right_half} & \primtypesig{grid -> grid} & Crop the right half of the grid (floor division)\\
\primname{top_half} & \primtypesig{grid -> grid} & Crop the top half of the grid (floor division) \\
\primname{bottom_half} & \primtypesig{grid -> grid}  & Crop the bottom half of the grid (floor division)\\

\midrule
\textbf{Uncropping}\\
\primname{repeatX} & \primtypesig{grid -> grid} & Stack two copies of grid horizontally\\
\primname{repeatY} & \primtypesig{grid -> grid} & Stack two copies of grid vertically\\
\primname{mirrorX} & \primtypesig{grid -> grid} & Horizontally mirror grid [abc]->[abccba]\\
\primname{mirrorY} & \primtypesig{grid -> grid} & Vertically mirror grid\\
\primname{ic_embed} & \primtypesig{grid -> grid -> grid} & Embeds a grid into a larger hull defined by a 2nd argument (zero-padded)\\

\midrule
\textbf{Colour manipulation}\\
\primname{topcol} & \primtypesig{grid -> colour} & The most common non-black colour\\
\primname{rarestcol} & \primtypesig{grid -> colour} & The least common non-black colour\\
\primname{ic_filtercol} & \primtypesig{colour -> grid -> grid} & Retains only pixels with the specified colour\\
\primname{ic_erasecol} & \primtypesig{colour -> grid -> grid} & Removes any pixels with the specified colour\\
\primname{setcol} & \primtypesig{colour -> grid -> grid} & Set all non-black pixels to the specified colour\\
\primname{set_bg} & \primtypesig{colour -> grid -> grid} & Set black pixels to the specified colour\\
\primname{get_bg} & \primtypesig{colour -> grid -> grid} & Return grid of background pixels in grid in specified colour\\
\primname{ic_invert} & \primtypesig{grid -> grid} & Replaces black with the \code{topcol}, replaces colours with black\\
\primname{colourHull} & \primtypesig{colour -> grid -> grid} & Set every pixel to a colour \\

\midrule
\textbf{Position manipulation}\\
\primname{getpos} & \primtypesig{grid -> pos} & Get the position of a cropped grid (default 0,0)\\
\primname{getsize} & \primtypesig{grid -> size} & Get the size of the grid\\
\primname{ic_toorigin} & \primtypesig{grid -> grid} & Reset a grid's position to $(0, 0)$\\

\midrule
\textbf{Morphology}\\
\primname{fillobj} & \primtypesig{colour -> grid -> grid} & Fill each closed object's interior with a specified colour\\
\primname{ic_fill} & \primtypesig{grid -> grid} & \code{fillobj} coloured blue\\
\primname{ic_interior} & \primtypesig{grid -> grid} & Return interior of closed objects only, coloured \code{topcol}\\
\primname{ic_center} & \primtypesig{grid -> grid} & Create a grid of $w/2, h/2$ with centred position, coloured blue\\
\primname{ic_makeborder} & \primtypesig{grid -> grid} & Draw border around objects image in blue (border only) \\
\primname{ic_spread} & \primtypesig{grid -> grid} & Each black cell is coloured with its closest neighbour colour \\
\primname{ic_spread_minor} & \primtypesig{grid -> grid} & \ocaml{ic_spread} ignoring the most common colour. \\

\midrule
\textbf{Counting}\\
\primname{countPixels} & \primtypesig{grid -> count} & Return the number of non-black pixels in the input grid\\
\primname{countColours} & \primtypesig{grid -> count} & The number of non-black colours in the input grid\\
\primname{countComponents} & \primtypesig{grid -> count} & The number of 4-connected objects in the image\\
\primname{countToXY} & \primtypesig{count -> colour -> grid} & Draw a new grid of $n \times n$ with the specified colour\\
\primname{countToX} & \primtypesig{count -> colour -> grid} & Draw a new grid of $n \times 1$ with the specified colour\\
\primname{countToY} & \primtypesig{count -> colour -> grid} & Draw a new grid of $1 \times n$ with the specified colour \\

\midrule
\textbf{Compression}\\
\primname{ic_compress2} & \primtypesig{grid -> grid} & Remove rows/columns which are duplicates of preceeding rows/cols\\
\primname{ic_compress3} & \primtypesig{grid -> grid} & Remove any entirely black rows/columns\\

\midrule

\end{tabular}

\begin{tabular}{lll}
\toprule
\textbf{Primitive} & \textbf{Type} & \textbf{Description}\\
\midrule
\textbf{Drawing}\\
\primname{ic_connectX} & \primtypesig{grid -> grid} & Join up any objects of the same colour horizontally\\
\primname{ic_connectY} & \primtypesig{grid -> grid} & Join up any objects of the same colour vertically\\
\primname{ic_connectXY} & \primtypesig{grid -> grid} & Connect in both X and Y\\
\midrule
\textbf{List creation}\\
\primname{ic_splitcols} & \primtypesig{grid -> list(grid)} & Split a grid based on colours\\
\primname{ic_splitall} & \primtypesig{grid -> list(grid)} & Split grid based on 4-connected objects\\
\primname{split8} & \primtypesig{grid -> list(grid)} & Split grid based on 8-connected objects\\
\primname{ic_splitcolumns} & \primtypesig{grid -> list(grid)} & Create $1\times n$ grids per column \\
\primname{ic_splitrows} & \primtypesig{grid -> list(grid)} & Create $n \times 1$ grids per row\\

\midrule
\textbf{List reduction}\\
\primname{pickcommon} & \primtypesig{list(grid) -> grid} & If there are repeated grids, return the most common\\
\primname{ic_pickunique} & \primtypesig{list(grid) -> grid} & If there is one unique grid, return it\\
\primname{pickmax_count} & \primtypesig{list(grid) -> grid} & Return grid with the most coloured cells\\
\primname{pickmax_neg_count} & \primtypesig{list(grid) -> grid} & Return grid with fewest coloured cells\\
\primname{pickmax_size} & \primtypesig{list(grid) -> grid} & Return grid with largest area\\
\primname{pickmax_neg_size} & \primtypesig{list(grid) -> grid} & Return grid with smallest area\\
\primname{pickmax_cols} & \primtypesig{list(grid) -> grid} & Return grid with most colours\\
\primname{pickmax_interior_count} & \primtypesig{list(grid) -> grid} & Return the grid with the most empty interior holes\\
\primname{pickmax_neg_interior_count} & \primtypesig{list(grid) -> grid} & Return the grid with the fewest empty interior holes\\
\primname{pickmax_x_pos} & \primtypesig{list(grid) -> grid} & Return right-most grid\\
\primname{pickmax_x_neg} & \primtypesig{list(grid) -> grid} & Return left-most grid\\
\primname{pickmax_y_pos} & \primtypesig{list(grid) -> grid} & Return uppermost grid\\
\primname{pickmax_y_neg} & \primtypesig{list(grid) -> grid} & Return lowermost grid\\

\midrule
\textbf{List processing}\\
\primname{mklist} & \primtypesig{grid -> grid -> list(grid)} & Initialise list from two elements\\
\primname{lcons} & \primtypesig{grid -> list(grid) -> list(grid)} & List cons\\

\midrule
\textbf{Composition}\\
\primname{ic_composegrowing} & \primtypesig{list(grid) -> grid} & Overlay grids from largest to smallest, taking into account position \\
\primname{overlay} & \primtypesig{grid -> grid -> grid} & Overlay two grids transparently. If same size, ignore position\\
\primname{logical_and} & \primtypesig{grid -> grid -> grid} & Pixel-wise AND between two grids. Uses colour of first grid\\

\midrule
\textbf{Higher-order functions}\\
\primname{mapSplit8} & \primtypesig{(grid -> grid) -> grid -> grid} & Apply a \primtypesig{(grid -> grid)} lambda to all objects individually\\

\midrule
\textbf{Gravity} \\
\primname{gravity_down} & \primtypesig{grid -> grid} & Move all objects down with gravity and collisions\\
\primname{gravity_up} & \primtypesig{grid -> grid} & Move all objects up with gravity and collisions\\
\primname{gravity_left} & \primtypesig{grid -> grid} & Move all objects left with gravity and collisions\\
\primname{gravity_right} & \primtypesig{grid -> grid} & Move all objects right with gravity and collisions\\
\bottomrule
\end{tabular}
}
\normalfont

\end{document}